\documentclass[preprint,12pt,3p]{elsarticle}

\usepackage{graphicx}               % Necessary to use \scalebox
\usepackage[utf8]{inputenc}
\usepackage[T1]{fontenc}

\usepackage{soul}
\usepackage{lipsum}
\usepackage{hyperref}
\usepackage[english]{babel} %Language English
\usepackage{url, graphicx, subcaption, makecell, pbox,amsmath, xcolor}
\usepackage{array}

\usepackage{lineno}

\begin{document}

\begin{frontmatter}

%\title{Segmenting overlapped cell clusters in biomedical images by concave point detection}

\title{Improving concave point detection to better segment overlapped objects in images}

%Segmenting overlapped objects in images  by concave point detection. A study to support the diagnosis of sickle cell disease.}

\author[1]{Miquel Miró-Nicolau}
\ead{miquel.miro@uib.es}
\author[1]{Gabriel Moyà-Alcover\corref{cor1}}
\ead{gabriel.moya@uib.es}
\author[2,3]{Manuel Gonz{\'a}lez-Hidalgo}
\ead{manuel.gonzalez@uib.es}
\author[1]{Antoni Jaume-i-Capó}
\ead{antoni.jaume@uib.es}

\cortext[cor1]{Corresponding author}
\address[1]{UGiVIA Research Group, University of the Balearic Islands, Dpt. of Mathematics and ComputerScience, 07122 Palma, Spain}
\address[2]{SCOPIA Research Group, University of the Balearic Islands, Dpt. of Mathematics and ComputerScience, 07122 Palma, Spain}
\address[3]{Health Research Institute of the Balearic Islands (IdISBa), E-07010, Palma, Spain}

% \thispagestyle{empty}

% \maketitle
% \thispagestyle{empty}

\begin{abstract}
%In this article we proposed a new method to separate these objects based on the detection of concave points as the local extreme of a contour curvature.

%Proposta

This paper presents a method {that improve state-of-the-art of the concave point detection methods} as a first step to segment overlapping objects on images. It is based on the analysis of the curvature of the objects' contour. The method has three main steps. First, we pre-process the original image to obtain the value of the curvature on each contour point. Second, we select regions with higher curvature and we apply a recursive algorithm to refine the previous selected regions. Finally, we obtain a concave point from each region based on the analysis of the relative position of their neighbourhood
%Given an image of an object cluster we compute the curvature on each point of its contour. Then, we select regions with the highest probability to contain an interest point, that is, regions with higher curvature. Finally we obtain an interest point from each region and we classify them between convex and concave.

%Presentam un mètode que millor la detecció de punts concaus, mitjançant un algorisme de detecció de punts d'interés a partir de la curvatura de l'objecte.

We experimentally demonstrated that a better concave points detection implies a better cluster division.
%Dataset
In order to evaluate the quality of the concave point detection algorithm, we constructed a synthetic dataset to simulate overlapping objects, providing the position of the concave points as a ground truth. 
%Segmentació cel·lules
As a case study, the performance of a well-known application is evaluated, such as the splitting of overlapped cells in images of  peripheral  blood  smears  samples  of patients with sickle cell anaemia. We used the proposed method to detect the concave points in clusters of cells and then we separate this clusters by ellipse fitting. 
%Resultats: Els resultats mostren que a millor detecció de punts concaus tenim una millor performance => no sé si és correcte dir aixo
 \end{abstract}


\end{frontmatter}

\newcommand{\miace}{OverArt\space}
\newcommand{\real}{ErythrocitesIDB2\space}

\section{Introduction} \label{introduccio}

Segmenting overlapped objects on images is a process that can be used as a first step in many biomedical and industrial applications, from blood cells study~\cite{gonzalez2014red} to grain analysis~\cite{yan2011new}. 
%In further steps, to analyse the information available in the images, it is necessary to obtain the full contours of the segmented objects.

In these applications an analysis of individual objects by their singular features is typically required. The existence of image areas with overlapping objects reduce the available information for some of the objects in the image, the so-called occlusion zones. These occlusion zones introduce an enormous complexity into the segmentation process and makes it a challenging problem that can be solved from multiple points of view and is still open.

Multiple strategies have been used to address this problem. Watershed algorithm~\cite{wahlby2004combining,rodriguez2005new,mosaliganti2012acme} and level sets~\cite{chang2007segmentation} are widely used on scenarios with well-defined boundaries between the overlapped objects. However, these methods are unable to separate the overlapped objects with an homogeneous intensity values  because the detection of initial markers is not accurate, provoking a diffuse boundary.   

To avoid these problems, another type of segmentation methods that can be used are those based on the detection of concave points. These points denote the positions where the contours of the different objects overlap and, at the same time, are the locations where the overlapped object changes from one of its sub-objects to another one. Once the concave points are detected, multiple techniques can be used to divide these objects. The advantages of the detection of concave points is that those are invariant to scale, color, rotation and orientation. 

Furthermore, the methods based on the detection of concave points can perform a good segmentation without using a big dataset and without input image size constraints unlike deep learning methods. Moreover, the detection of concave points for segmenting overlapped objects can be considered to be transparent because presents simulatability, decomposability and algorithmic transparency~\cite{arrieta2020explainable}. %To trust the behavior of intelligent systems, especially in health, they must be able to explain their results to the experts.

The better concave points detection, the better cluster division. Therefore, we propose a new method that increases the precision of the concave point detection with respect to state-of-the-art. This method is based on the analysis of the contour's curvature. To evaluate our proposal, we constructed a synthetic dataset to simulate overlapping objects and providing the position of the concave points as a ground-truth. We used this dataset to compare the detection capacity  and the spatial precision of the proposed method with the state-of-the-art. As a case study, we evaluated the proposed concave point detector with a well-known application, such as the splitting of overlapped cells in microscopic images of peripheral blood smear samples of red blood cells (RBC) of patients with sickle-cell disease (SCD). It is important to point out that this algorithm is not limited to cell segmentation, it can be used for other applications where separation between overlapping objects 
is required.

The remainder of this paper is organized as follows. In the next section we describe the related work. In Section~\ref{methodology} we explain the proposed method for the efficient detection of concave points. In Section~\ref{experimental}, we specify the experimental environment and we describe the datasets used for experimentation. Section~\ref{results} is devoted to discuss the results and comparison experiments  obtained after applying the proposed method to synthetic and real images of clusters of objects. Finally, in Section~\ref{conclusion} we give the conclusions of our work.

\section{Related Work}
\label{sec:related_work}

In the state-of-the-art we can find multiple approaches based on concave point detection. Following the taxonomy proposed by Zafari \textit{et. al.}~\cite{zafari2017comparison}, but adding a category for other methods, we classified these methods in four categories: skeleton, chord, polygon approximation, curvature and others.

\subsection{Skeleton}

The approaches based on skeleton use the information of the boundary and its medial axis to detect concave points. Song and Wang ~\cite{song2009new} identified the concave points as local minimums of the distance between the boundary of the object and its medial axis. The medial axis was obtained by an iterative algorithm based on the binary thinning algorithm followed by a pruning method. Samma \emph{et al.}~\cite{samma2010combining} found the concave points by intersecting the boundary of the object, obtained by applying the morphological gradient and the skeleton of the dilated background image.

%These methods uses different algorithms to find the skeleton. 
These methods need a big change in the curvature to detect the existing concave points, so, skeleton-based methods tends to fail on contours objects with smooth curvatures.

\subsection{Chord}

Chord methods use the boundary of the convex hull area of the overlapped objects.  This boundary consists of a finite union of curve segments, each of one is an arc of the object's contour $C$, or it is a  line  with its end points on $C$, named a chord.  If $L_i$ is one of those lines (a chord) and $C_i$ is the segment of C with its same end points, the union of $L_i$ with $C_i$ generate a simple closed curve, which determines a convexity defect. Chord methods use the convex hull contour and the convexity defects to detect concavities. The main idea of these approaches is to identify the furthest points between the contour and the convexity defect. 

Multiple solutions used the chord analysis to extract convex points. Farhan \emph{et al.}~\cite{farhan2013novel} proposed a method to obtain the concave points by evaluating the line fitted to the contour points, a concave point is detected if the line that joins the two contour points does not reside inside the object. The proposal of Kumar \emph{et al.}~\cite{kumar2006rule} used the boundaries of concave regions and their  corresponding  convex  hull chords. The concave points were defined as the points on the boundaries of concave regions that maximize perpendicular distance from the convex hull chord. Similarly, Yeo \emph{et al.}~\cite{yeo1994clump} and LaTorre \emph{et al.}~\cite{latorre2013segmentation} proposals applied multiple constraints to the area between the convexity defect and the contour to determine its quality. 

Chord methods consider that only exists one concave point for each convexity defect, in clusters with more than two objects this assumption is not always true, therefore some concave points are missdetected. %The drawback of these approaches is the misdetection of some concave points.% between a chord and the curvature. 

\subsection{Polygon Approximation}

Polygon approximation is a set of methods that represents the contours of the objects through a sequence of dominant points. These methods aims to remove noise approximating the contour to a simpler object. 

Bai \emph{et al.}~\cite{bai2009splitting} developed a brand new algorithm to follow this approximation. Their algorithm analysed the difference between a set of contour points and the straight line that connected their extremes. The points with a big distance to this previously defined line were considered dominant points. Chaves \emph{et al.}~\cite{chaves2015concave} used the well-known algorithm of Ramer–Douglas–Peucker (RDP) \cite{douglas1973algorithms} to approximate the contour and the concave points were detected using conditions on the gradient direction at the contour. %Lines joining concave points are candidates to split lines, that are discriminated using a set of restrictions.
Another similar approach to detect these points was presented in \cite{zafari2020resolving} by Zafari \emph{et al.}  where the authors proposed a parameter free concave point detection to extract dominant points. They selected the concave ones using a condition based on the scalar product of consecutive dominant points described in \cite{zhang2012method}.

Same authors~\cite{zafari2015segmentation}, used a modified version of curvature scale space proposed in \cite{he2004curvature} to find interest points. Finally they discriminated them between concave and convex points. %Instead of the rest of the other methods they do not calculate the angular change, all the dominant points of the CSS are considered to be interest points.

Zhang \emph{et al.}~\cite{zhang2020structure}, similarly to \cite{zafari2015segmentation},  used the modified version of curvature scale space to make the object approximation. From the application of the previously described object approximation algorithms they obtained a set of dominant points, among them the concave points were found. The concave points were detected by evaluating angular change on these dominant points, and these angular changes were evaluated using the \textit{arctangent} criteria. The points with an angular change higher than a threshold were classified as concave points.

These methods are highly parametric and are not robust to object size changes. Another weakness of these methods is that they deform the original silhouette to simplify it. The approximation is a trade-off between the lack of precision in the position of the concave points and the smooth applied to the contour. This trade-off affects the final results.

\subsection{Curvature}

The methods that falls in this category identifies the concave points as the local extreme of the curvature. The curvature, $\kappa$, at each point $q_i=(x_i,y_i)$ of the contour is computed as: 

\begin{equation}
    \kappa (q_i) = \frac{x^{'}_i \cdot y''_i - y'_{i} \cdot x^{''}_{i}}{(x^{'2}_i~+~y^{'2}_i)^{3/2}},
    \label{eq:curvature}
\end{equation}
where $x_i$ and $y_i$ are the coordinates of the contour points.

%Multiple methods are used to approximate the value of $\kappa$. 
Wen \emph{et al.} \cite{wen2009delaunay} calculated the derivative by convolving the boundary with Gaussian derivatives. Gonz{\'a}lez-Hidalgo \emph{et al.}~\cite{gonzalez2014red} used the \emph{k}-curvature and the \emph{k}-slope to approximate the value of the curvature. The dominant points, the ones with most curvature, can be located in both concave and convex regions of the contours. In~\cite{zafari2017comparison} three different heuristics are described to detect the concave ones.

These methods tend to fail when multiple concave points are located in small areas, the main reason for this error is the loss of precision by the approximation of the curvature value. Another source of problems is the existence of noise, these algorithms tend to identify the noise on the contour as changes on the curvature, one way to fix this problem is to use a more coarse approximation.

\subsection{Other methods}

Despite the taxonomy proposed by Zafari \emph{et al.}~\cite{zafari2017comparison} there are other techniques to find concave points that do not fall into any of the previous categories. We describe some of these works below. 

Fern{\'a}ndez \emph{et al.}~\cite{fernandez1995new} defined a sliding window for the contour and calculates the proportion of pixels that belongs to the object and the pixels that belong to the background on this window. This proportion determined the likelihood of an existing concavity on the evaluated point. He \emph{et al.}~\cite{he2014automated} adapted this method to use it in three dimensions. Best results were obtained in scenarios with high concavity. This method is very sensitive to changes on the size of the objects and its accuracy decreased with the existence of noise. These two problems are a consequence of the lack of generalization of the method.

Wang \emph{et al.}~\cite{wang2012clump} proposed a bottleneck detector. They defined a bottleneck as a set of two points that minimize its euclidean distance and maximize the distance on the contour. The set of points that defined a bottleneck were the concave points. A cluster can contain multiple bottlenecks. This algorithm was unable to discover how many elements belong to a cluster, the number of elements is an hyperparameter of the algorithm. Another limitation was that they did not considered clusters with an odd number of concave points.

Zhang and Li~\cite{zhang2017automated} proposed a method to find the concave points with a two step algorithm. First, they detected a set of candidates points with the well known Harris corner detector~\cite{harris1988combined}. Second, they selected the concave points with two different algorithms, one for obvious concave points and another for uncertain concave points. Their algorithm have a high number of parameters. This higher number of parameters have two different consequences: on one hand the method is highly adjustable to the features of the overlapped objects, on the other hand this amount of parameters provoke a high complexity of the algorithm. %inducing the failure to deal with objects of different size.

\section{Methodology} \label{methodology}

\begin{figure*}[!htb]
\begin{center}
	\includegraphics[width = \linewidth]{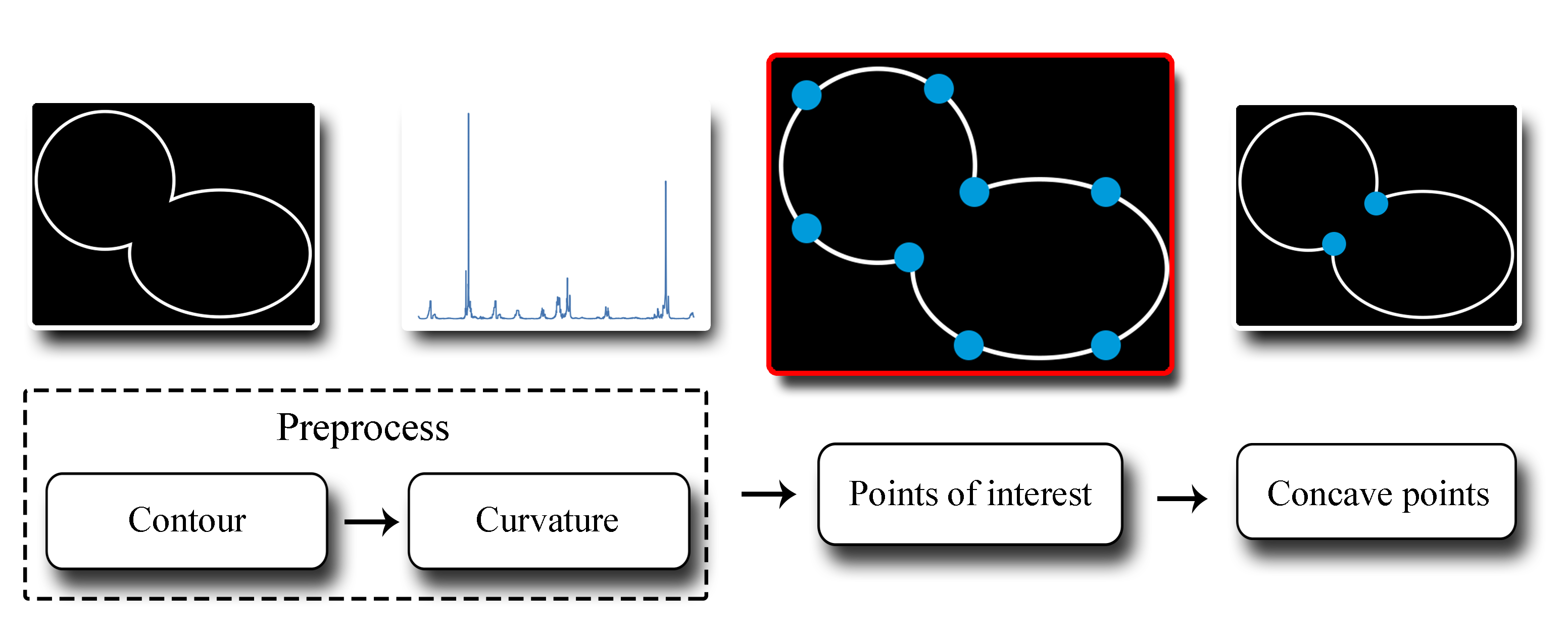}
\end{center}
\caption{Flow chart of the proposed method to obtain concave points from a contour.}\label{fig:flow_method}
\end{figure*}

Motivated by the state-of-the-art performance and with the aim of improving the results on the challenging task of separating overlapped objects, we propose a new method based on Gonz{\'a}lez-Hidalgo \emph{et al.}~\cite{gonzalez2014red}  to find concave points. The whole process is summarized in Figure \ref{fig:flow_method}.

\subsection{Pre-procesing}

%We define a contour as a set of $n$ ordered points with orientation that defines the shape of an object, $c=\{ (x_0, y_0), (x_1, y_1), \ldots, (x_{n-1}, y_{n-1}) \}$. 

To obtain objects from images we applied a segmentation algorithm. Depending on the complexity of the data we face, we can use different techniques: from Otsu~\cite{otsu1979threshold}, a simpler, non parametric, thresholding technique, to a more complex approach as Chan-Vese~\cite{chan1999active} that needs to be parametrised. Once the object is segmented we obtained its contour and we removed the noise generated by the segmentation technique using the RDP algorithm by \cite{douglas1973algorithms}. Finally, we computed the curvature on each point through a well-known technique, the \emph{k}-curvature \cite{pavlidis1980algorithms}. This technique considers the curvature of every point as the difference of its slope. The \emph{k}-curvature is separable, it allowed us to made the calculation independently for each direction.

\subsection{Points of interest}

In this section we define a methodology to find points of interest by analyzing the curvature of each contour point, a point of interest can be either a concave or convex point, in next step we need to identify the concave ones.%The input of this algorithm is the list of contour points with the values of its curvatures.  

The first step to obtain the interest points was to determine subsets of contiguous points that included those with highest curvature. We called these subsets, regions of interest. All points that belong to a region of interest have a curvature greater than a certain threshold and it is defined by its start and end points. The process to determine the regions of interest is a recursive procedure that we describe below, which provided a list of regions of interest and ensured the presence of a point of interest within each of them.

Let $C$ and $t$ be the set of contour points and an initial threshold for the curvature value, respectively. We also defined two other thresholds to control the length of the regions of interest, namely $ l_{min} $ and $ l_{max} $.  The  $l_{min} $ value aims to avoid having an excessive number of regions and reduce the effect of noise on the object contour. The $ l_{max} $ value is useful to prevent that the point of interest is located in an excessively large region, as we were interested to extract one concave point from each region. The recursive procedure that allowed us to detect the regions of interest is as follow:
\vskip 0.3 cm

\begin{figure}[!htpb]
    \centering
    \begin{subfigure}[t]{0.20\linewidth}
        \centering
        \captionsetup{width=.9\linewidth}
        \includegraphics[width=0.98\textwidth]{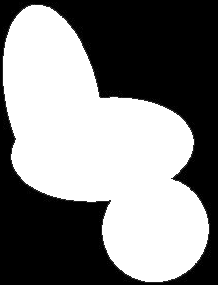}
        \caption{}\label{fig:succesive_steps_a}
    \end{subfigure}%   
    %\hspace{0.3cm}
    \begin{subfigure}[t]{0.20\linewidth}
        \centering
        \captionsetup{width=.9\linewidth}
        \includegraphics[width=0.98\textwidth]{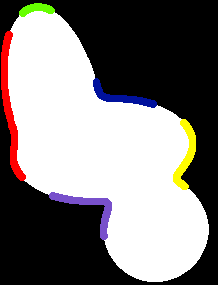}
        \caption{}\label{fig:succesive_steps_b}
    \end{subfigure}%   
    % \hspace{0.3cm}
    \begin{subfigure}[t]{0.20\linewidth}
        \centering
        \captionsetup{width=.9\linewidth}
        \includegraphics[width=0.98\textwidth]{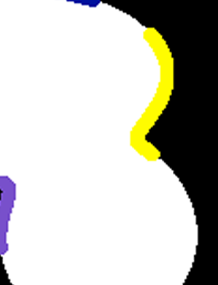}
        \caption{}\label{fig:succesive_steps_d}
    \end{subfigure}%   
        \begin{subfigure}[t]{0.20\linewidth}
        \centering
        \captionsetup{width=.9\linewidth}
        \includegraphics[width=0.98\textwidth]{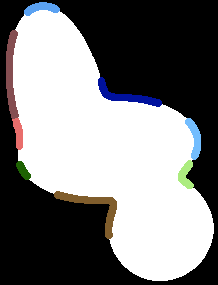}
        \caption{}\label{fig:succesive_steps_c}
    \end{subfigure}%   
        \begin{subfigure}[t]{0.2\linewidth}
        \centering
        \captionsetup{width=.9\linewidth}
        \includegraphics[width=0.98\textwidth]{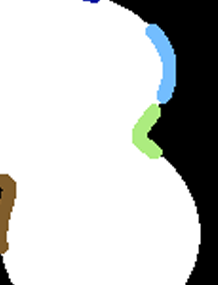}
        \caption{ }\label{fig:succesive_steps_e}
    \end{subfigure}%   
    \caption{Successive steps to identify regions of interest. (a) Original image. (b) Regions of interest detected in Step 1. Different regions of interest are marked with different colors. (c) Detail of a region of interest detected in Step 1. (d) Regions of interest detected by the recursive procedure, Step 2. (e) Detail of Step 2. }%with a point of interest within each of them.}
    \label{fig:succesive_steps}
\end{figure}

\noindent{\bf Step 1:} Taking the contour points $C$, we constructed a list of regions of interest \texttt{l\_regions} selecting all non-overlapping sets of adjacent points which curvature was greater than $t$. If necessary we updated $t$. The original image is displayed in Figure \ref{fig:succesive_steps_a}. Figure \ref{fig:succesive_steps_b} depicts the list of regions of interest detected, different regions of interest are marked with different colors.

\noindent{\bf Step 2:} Let $r$ be a region of interest in \texttt{l\_regions}, we denote by $\ell (r)$ its length. We considered three possible cases:
\begin{itemize}
\item{Case 1:} If $l_{min} \leq \ell(r) \leq l_{max}$, we continued with the next region of interest. 
\item{Case 2:} If $\ell(r)  >  l_{max}$, we returned to Step 1 with $r$ and $t+\delta t$. We updated {\tt l\_regions} with the regions into which $r$ is divided, and $r$ is removed from the list.
\item{Case 3:} If $\ell(r)  < l_{min}$,  we combined the region $r$ with its closest region. That is, we looked for $r_{closest} $ such that $ d (r, r_{closest}) <k $, where $ k $ was the displacement allowed to calculate the $ k $-curvature. Let $r_{new}=r\cup r_{closest}$ be the new region. 
\begin{itemize}
\item[a)]	If $\ell (r_{new}) >   l_{min}$, we returned to Step 1 with $ r_{new}$ and $t+\delta t$. If necessary updated {\tt l\_regions}.
\item[b)]	If $\ell (r_{new}) <    l_{min}$, we moved to Case 3 with $ r_{new}$ instead $r$. 
\end{itemize}
\end{itemize}

\vskip 0.3 cm 
The process ended when all the regions of interest in {\tt l\_regions} had a length between $ l_{min} $ and $ l_{max} $.  In Figure \ref{fig:succesive_steps_c} we display the final list of regions of interest we detected. As above, different regions of interest are marked with different colors. Moreover, in Figure \ref{fig:succesive_steps_d} we show a zoom with an initial region of interest, and in Figure \ref{fig:succesive_steps_e} we depict its final state.  As we can see, the initial region of interest is divided into two new regions, each containing a point of interest.

After this recursive procedure, we obtained a list of regions of interest, where each of the regions contained a point of interest, that is a concave or a convex point. Finally, we identified one interest point inside each region of interest. We used the weighted median of the curvature to locate them, because it is a well-known technique that assume that the point of interest is located near the center of the region but this central position is not a perfect location and it can be improved.

\subsection{Concave points selection}

\begin{figure*}[!h]
\centering
 \begin{subfigure}[b]{0.31\linewidth}
    \centering
    \includegraphics[width=0.90\textwidth]{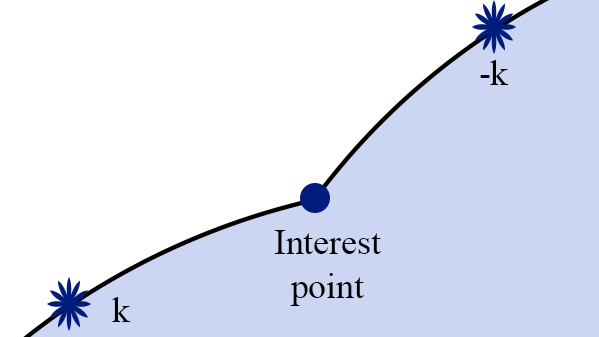}
    \caption{We select two points at k and -k distance from the interest point.} \label{fig:concavity_first}
 \end{subfigure}\hfill 
 \begin{subfigure}[b]{0.31\linewidth}
    \centering
    \includegraphics[width=0.90\textwidth]{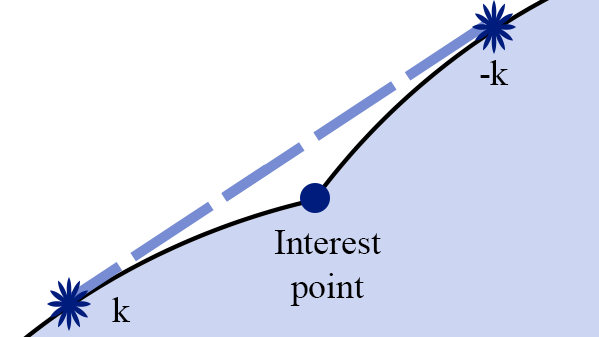}
    \caption{We determine the straight line between previously selected points.} \label{fig:concavity_second}
 \end{subfigure}\hfill 
 \begin{subfigure}[b]{0.31\linewidth}
    \center
    \includegraphics[width=0.90\textwidth]{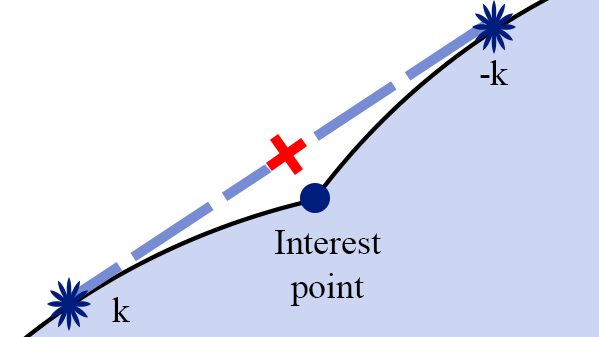}
    \caption{The interest point is classified concave if the red point is outside the object.} \label{fig:concavity_third}
 \end{subfigure}\hfill 
 \caption{Steps to discriminate between concave and convex points.}
\end{figure*}

Once we determined the set of points of interest we needed to identify the concave ones. This part of the algorithm was based on the analysis of the relative position of the neighbourhood of each point~\cite{gonzalez2014red}. The classification phase followed three steps: 
\begin{enumerate}
    \item \textbf{Determine two $k$-neighbour points}: We selected two points on the contour, they were located at $k$ and $-k$ distance relative to the interest point. This step is described in Figure \ref{fig:concavity_first}.
    \item \textbf{Definition of a line between the $k$-neighbours}: We built a straight line between the points selected in the previous step, see Figure \ref{fig:concavity_second}. 
    \item \textbf{Middle point of the line}: We classified the point  as concave if the middle point of the previously defined line was outside the object, otherwise was classified as convex. See Figure \ref{fig:concavity_third}.
\end{enumerate}

%The algorithm described in this section identifies the position of all concave points that belongs to a contour.% Is divided into three main steps: first, the calculation of the curvature for each point and the identification of sets of contiguous points with high values of curvature, regions of interest. Second, from this set of point obtained in the previous step we extract points of interest. Finally, we select the concave points. 

\section{Experimental Setup} \label{experimental}
% Explicar que el nostre objectiu és fer un bon detector de punts concaus i demostrar que si millora la detecció dels punts concaus millora la separació dels objectes 
The experimental setup is designed to evaluate the performance of the concave point detector compared to the state-of-the-art and to prove that a better concave point detection implies a better segmentation of overlapping objects. %In particular, in this section we describe the image datasets, the metrics we used to measure the performance, the methods selected to be compared and the two experiments we designed.

\subsection{Datasets}
\label{sec:datasets}
We used two different sets of images. On the one hand, we created a set of synthetic images, that we called \miace dataset. It contains 2000 images, each with 3 overlapping objects with annotated concave points as groundtruth. On the other hand, we used the \textit{ErythrocytesIDB2} dataset of real images from Gonz{\'a}lez-Hidalgo et al.~\cite{gonzalez2014red}, it contains 50 images of peripheral blood smears samples of patients with sickle cell anaemia. We used this dataset to check if  the spacial precision of the concave points detector method affects the results of the overlapping object segmentation.

\subsubsection{\miace Dataset}
\begin{figure}[!b]
\centering
 \begin{subfigure}[b]{0.33\linewidth}
    \centering
    \includegraphics[width=0.97\textwidth]{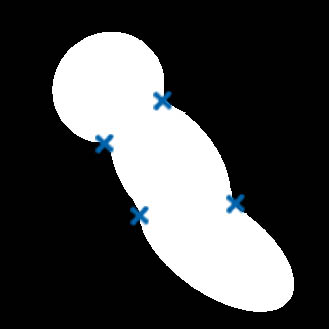}
 \end{subfigure}%   
  \begin{subfigure}[b]{0.33\linewidth}
    \centering
    \includegraphics[width=0.97\textwidth]{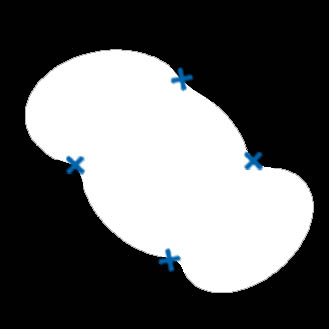}
 \end{subfigure}%   
   \begin{subfigure}[b]{0.33\linewidth}
    \centering
    \includegraphics[width=0.97\textwidth]{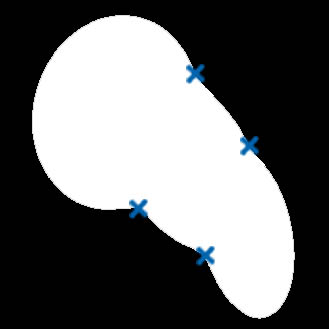}
 \end{subfigure}%   
  \caption{Three examples from the \miace dataset. In blue color the concave points, in white color the clusters defined by three overlapped ellipses.}\label{fig:second-dataset-example}
\end{figure}

We generated the \miace dataset in order to obtain a ground truth of the concave points on overlapped objects. Each image of the dataset contains a cluster with three overlapped ellipses. We put three ellipses in order to simulate the real case of the red blood cells in microscopic images.
%a good trade-off between complexity and reality.}  
The code is available at \url{https://github.com/expainingAI/overArt}. The same set of images we used in this paper can be created choosing number 42 as the random seed. 
%The proposed method to detect concave points is only necessary when there are at least two overlapped  objects. Nevertheless, we add another ellipse to increase the complexity of the problem. 
Figure \ref{fig:second-dataset-example} depicts three different examples of this dataset. 

The ellipses of each image were defined by three parameters: the rotation, the \textit{feret} diameter size and its center. We randomly generated these values by the set of constraints described in Table \ref{tab:parameters_generators}. %This constrains are completed with the next restriction: none of the three cells must be outside the cluster. 

To construct each cluster we located the first ellipse in the center of the image. The positions of the other two ellipses are related to this first one. We randomly selected the location of the second ellipse  inside the area defined by the minimum and maximum distance to the center of the first ellipse. Finally, we followed the same process with the third ellipse, it was randomly placed inside the area defined by the minimum and maximum distance to the center of the first and second ellipses. 

\begin{table}[!htb]
\centering
\begin{tabular}{lr}
\hline
Parameter                               & Value                     \\ \hline
Minimum \textit{feret}                  & $45$ px                        \\
Maximum \textit{feret}                  & $100$ px                       \\
Minimum distance between centers        & $45$ px                        \\
Maximum distance between centers        & $85$ px                        \\
Minimum rotation                        & $0$ º~~ \\
Maximum rotation                        & $360$ º~~ \\ \hline
\end{tabular}
\caption{Parameters we used to generate the \miace dataset. The table detail the range of the values that defines each ellipse. The distance metrics are defined in pixels (px). Angles are defined in degrees. }
\label{tab:parameters_generators}
\end{table}

To compare the performance of the precision of the different methods to find the concave points we needed a ground truth of its location. We calculated it and we added this information to the dataset. A concave point is defined by the position where two or more ellipses intersects and must be located over the contour that defines the overlapping region.

The overlapped objects are defined by the equation of the ellipse, see Eq.~(\ref{eq:ellipse_1}) and Eq.~(\ref{eq:ellipse_2}). For each image of the dataset we obtained the position of all of its concave points by analytically solving Eq.~(\ref{eq:ellipse_3}). 

\begin{equation}
    {\lambda}_1 = \frac { ((x-center_x)\cos(\phi)+(y-center_y) \space \sin(\phi))^2 }{ (a^2)},
    \label{eq:ellipse_1}
\end{equation}

\begin{equation}
    {\lambda}_2 = \frac{((x-center_x) \sin(\phi)-(y-center_y) \space \cos(\phi))^2}{(b^2) },
    \label{eq:ellipse_2}
\end{equation}

\begin{equation}
    {\lambda}_1 + {\lambda}_2  = 1,
    \label{eq:ellipse_3}
\end{equation}

\noindent where $x$ and $y$ are the unknown variables, $center_x, center_y$ defines the central point of the ellipse, $\phi$ the angle between the horizontal axis and the ellipse feret. Finally $a$ and $b$ represents each semi-axis.

\subsubsection{\real Dataset}
In this work we used microscopic images of blood smears, collected from  \textit{ErythrocytesIDB2} \cite{gonzalez2014red}, available at \url{http://erythrocytesidb.uib.es/}. The images consist of peripheral blood smears samples of patients with sickle cell anaemia classified by a specialist from ``Dr. Juan Bruno Zayas'' Hospital General in \textit{Santiago de Cuba}, Cuba. The specialist's criteria was used as an expert approach to validate the results of the classification methods. 

The patients with sickle-cell disease (SCD), are characterized by red blood cells (RBCs) with the shape of a sickle or half-moon instead of the smooth, circular shape as normal cells have. %WHO document "Global epidemiology of haemoglobin disorders and derived service indicators"~\cite{modell2008global} indicates that around a 5\% of the world's population carries trait genes for haemoglobin disorders, mainly sickle cell disease and thalassemia. The document also indicates that the percentage of people who carry these genes is as high as 25\% in some regions and over 300,000 babies with severe hemoglobin disorders are born each year. SCD is spread among people whose ancestors are from sub-Saharan Africa, India, Saudi Arabia and Mediterranean countries. 
In order to confirm the SCD diagnose, peripheral blood smear samples are analyzed by microscopy to check for the presence of the sickle-shaped erythrocytes and compare their frequency to normal red blood cells. 
The peripheral blood smear samples always include overlapped or clustered cells, and the sample preparation process can affect the quantity of overlapping erythrocytes in the images studied. Clinical laboratories typically prepare blood samples for microscopy analysis using the dragging technique, using this technique, more cell groups are apparent in the samples due to the spreading process~\cite{gonzalez2014red}.

\begin{figure}[!b]
\centering
 \begin{subfigure}[b]{0.45\linewidth}
    \centering
    \includegraphics[width=0.75\textwidth]{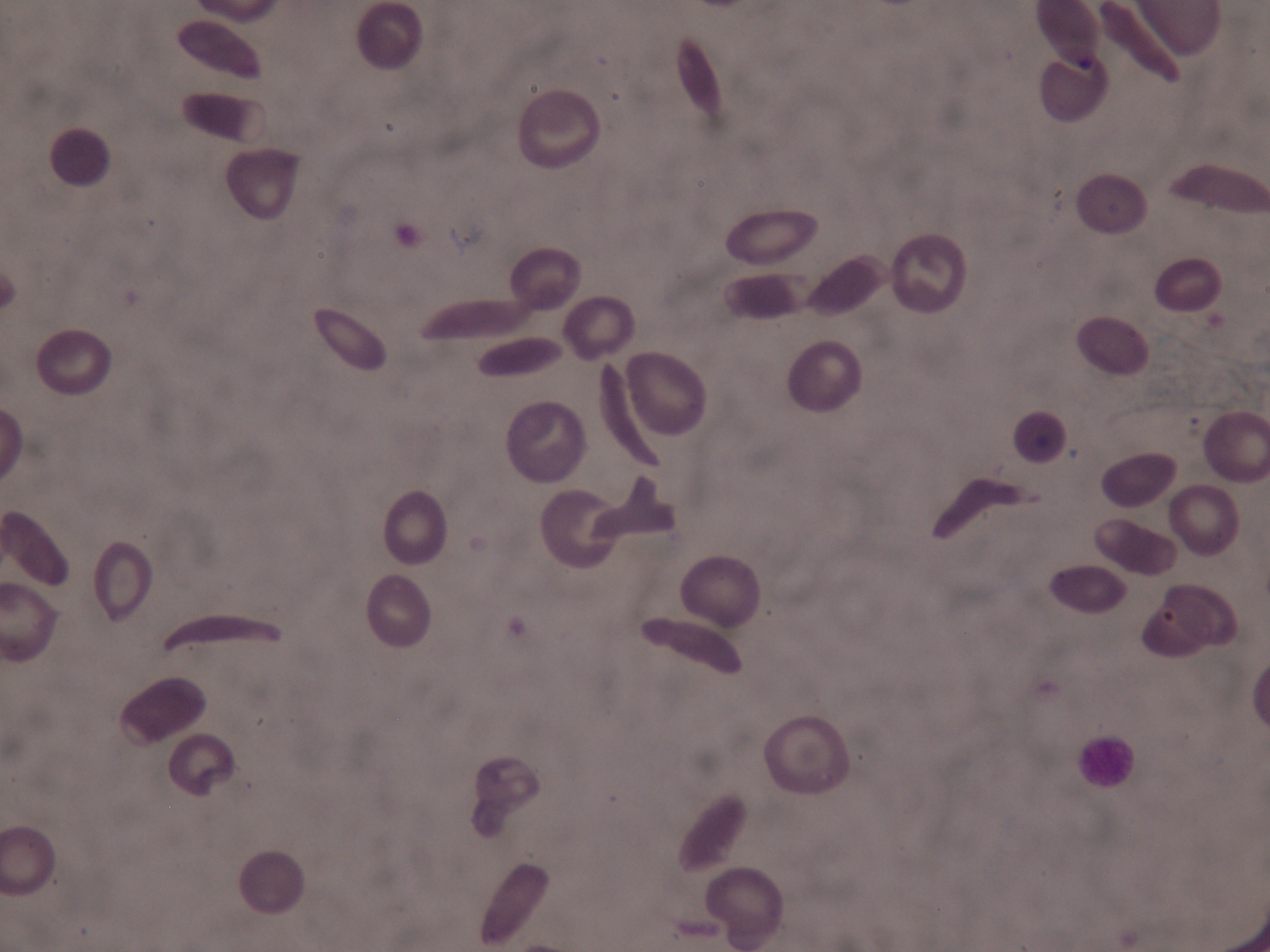}
    \caption{Example of image of the \real dataset.}
 \end{subfigure}%   
 \hspace{0.3cm}
 \begin{subfigure}[b]{0.45\linewidth}
    \centering
    \includegraphics[width=0.75\textwidth]{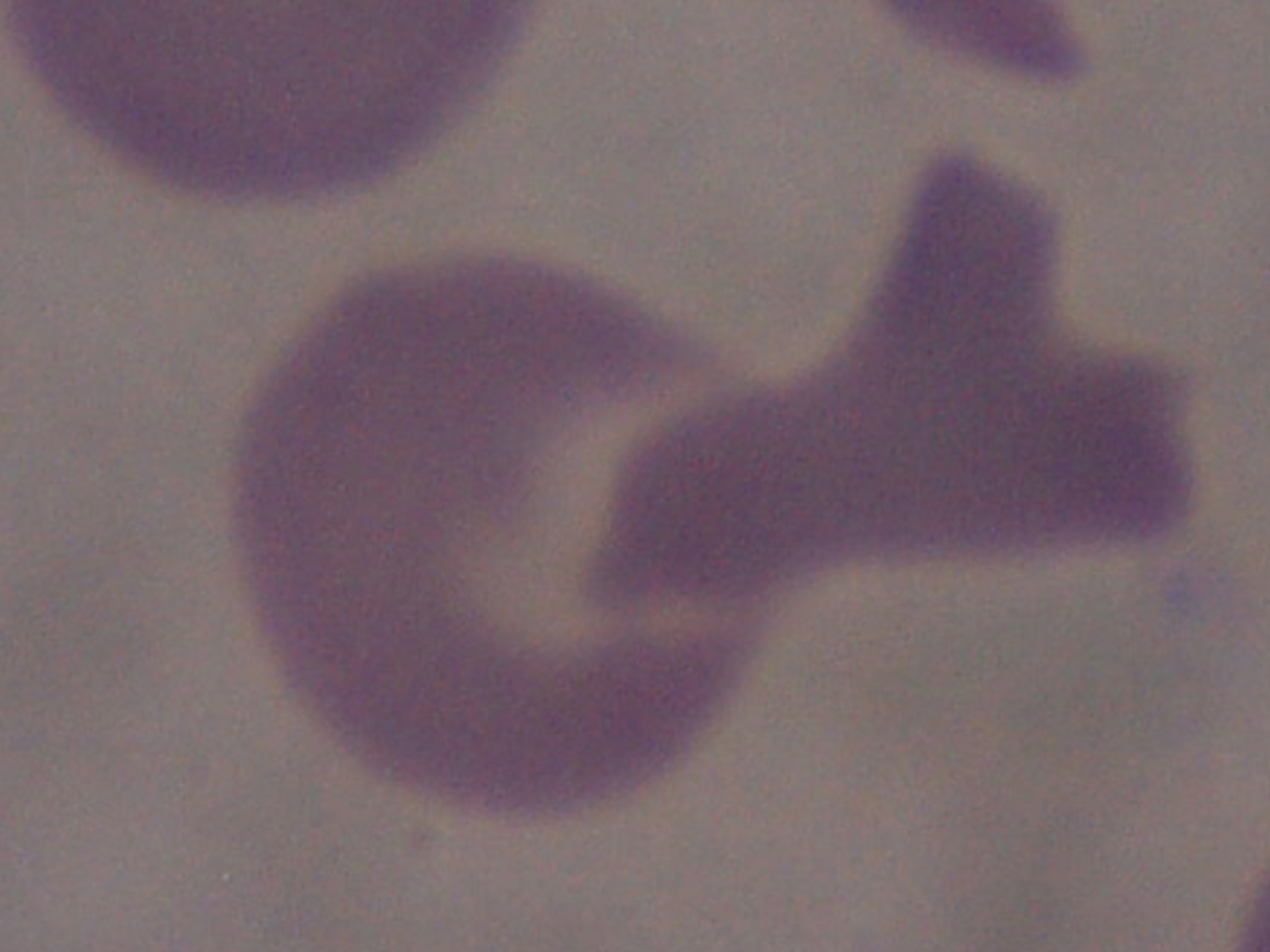}
    \caption{Detail of the image containing a cluster.}
 \end{subfigure}%
  \caption{Sample of patient with sickle cell anemia from \real dataset. }\label{fig:first-dataset-example}
\end{figure}

\begin{figure}[!hptb]
\centering
 \begin{subfigure}[t]{0.30\linewidth}
    \centering
    \includegraphics[width=1\textwidth]{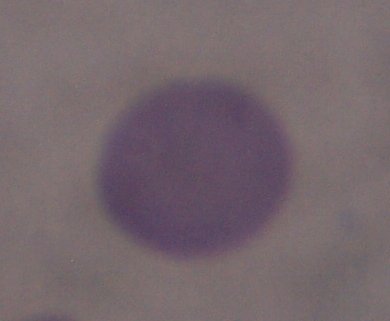}
    \caption{Example of healthy \textit{circular} cell.}
 \end{subfigure}%
  \hspace{0.3cm}
 \begin{subfigure}[t]{0.30\linewidth}
    \centering
    \includegraphics[width=1\textwidth]{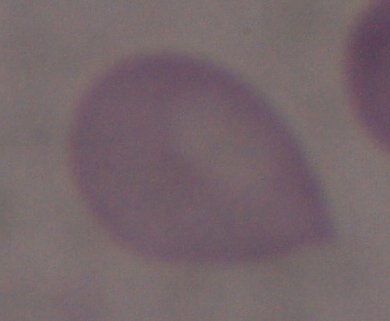}
    \caption{Example of \textit{other} cell.}
 \end{subfigure}%   
 \hspace{0.3cm}
  \begin{subfigure}[t]{0.30\linewidth}
    \centering
    \includegraphics[width=1\textwidth]{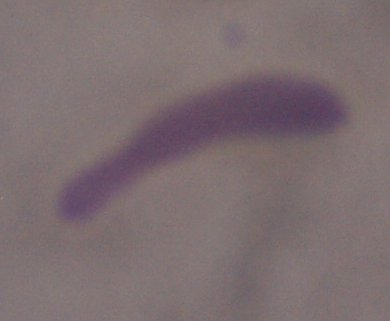}
    \caption{Example of \textit{elongated} cell.}
 \end{subfigure}%   
 \caption{Examples of the three types of cells present in the \real dataset. \textit{Elongated} cells are also known as sickle cells.}\label{fig:types_of_cell}
\end{figure}

Each image were labeled by the medical expert. There are 50 images with different number of cells (see Figure \ref{fig:first-dataset-example}), this set of images contains 2748 cells. These cells belongs to three classes defined by the medical experts. Those are circular, elongated and others as can be seen in figure \ref{fig:types_of_cell}.

% Performance measures-> revisar
\subsection{Performance measures}
\label{sec:performance_measures}
The results of our algorithm should be measured with multiple numerical and well defined metrics in order to ensure its quality. The objective of these metrics is to evaluate the precision on the prediction of the position of a concave point and how this change in precision affects on the split of overlapped objects. We used five different metrics: MED, F1-Score, SDS\_Score, MCC and CBA. %Two of these five metrics are more appropriate for unbalanced problems, since we are faced with a classification problem with unbalanced data.

\begin{itemize}
    \item \textbf{Mean of the Euclidean distance (MED).} Let  $f=\{C_i\}_{i=1}^{p}$ be the detected concave points for the proposed method for a given image, and  $GT=\{GC_i\}_{i=1}^{l}$ be the ground truth concave points of that image,  we know that it may be that  $l \neq p$.  However, for each point $GC_j$ exists $C_{i_j} \in f$ such that $d(GC_j , C_{i_j})$ is minimum, then  we define the $\mbox{MED}$ performance measure by Eq.~(\ref{eq:med}). 
    
    \item \textbf{F1-Score}. It is a standard and widely used measure. It is the harmonic mean of precision and  recall, see Eq.~(\ref{eq:f1_score}). The precision and the recall depends on the number of false positives (FP), true positives (TP) and false negatives (FN). We also included the precision and the recall to the results in order to explain the F1-Score.% The definition of these parameters depends on the nature of each experiment. 
    
    \item \textbf{Matthew’s Correlation Coefficient (MCC)}. Introduced in~\cite{matthews1975comparison}, is a correlation measure between the prediction and observation.  We used the adaptation proposed by Mosley \emph{et al.}~\cite{mosley2013balanced} for multi class problems, see Eq.~(\ref{eq:mcc}). This metric lies in [-1,1] range where -1 represents perfect missclasification, 1 a perfect classification and 0 a random classification. It is designed to deal with unbalanced data.
    
    \item \textbf{Class Balance Accuracy (CBA)}. Introduced by Mosley \emph{et al.}~\cite{mosley2013balanced}. Represents the overall accuracy measure built from an aggregation of individual class metrics. This measure is designed to deal with unbalanced data. See Eq.~(\ref{eq:CBA_score}).
    
    \item {\textbf{Sickle cell disease diagnosis support score (SDS-Score)}.} Proposed by Delgado-Font \emph{et al.}~\cite{delgado2020diagnosis}, the SDS-Score indicates the usefulness of the method for the diagnosis of patients with sickle cell disease. This metric does not consider as a mistake a misclassification between elongated and other cells (or vice versa), due to the nature of the disease. See Eq.~(\ref{eq:sds_score}).
\end{itemize}

\begin{equation}
    MED = \frac{\sum_{j=1}^{l} d(GC_j, C_{i_j})}{l},
    \label{eq:med}
\end{equation}

\begin{equation}
    Precision = \frac {TP}{ TP + FP },
    \label{eq:precision}
\end{equation}

\begin{equation}
    \centering
    Recall = \frac {TP}{ TP + FN },
    \label{eq:recall}
\end{equation}

\begin{equation}
    F1-Score = 2 \cdot \frac{Precision \cdot Recall}{Precision + Recall},
    \label{eq:f1_score}
\end{equation}

\begin{equation}
    MCC=\frac{\sum\limits_{i,l,m=1}^{z}c_{ii} \cdot c_{ml} - c_{li} \cdot c_{im}} {\sqrt{\sum\limits_{z=1}^{n}(\sum\limits_{l=1}^{n}c_{lz})(\sum\limits_{\substack{f,g=1 \\ f \neq z}}^{z}c_{gf})} \sqrt{\sum\limits_{l=1}^{z}(\sum\limits_{i=1}^{z}c_{il})(\sum\limits_{\substack{f,g=1 \\ f \neq l}}^{z}c_{fg})}},
    \label{eq:mcc}
\end{equation}
 
\begin{equation}
    CBA = \frac{1}{z} \cdot \sum\limits_{i=1}^{z}\frac{c_{ii}}{max(\sum\limits_{j=1}^{3} c_{ij}, \sum\limits_{j=1}^{3} c_{ji})},
    \label{eq:CBA_score}
\end{equation}

\begin{equation}
    SDS-Score = \frac{\sum\limits_{i=1}^{3}c_{ii} + c_{23}+c_{32}}{\sum\limits_{i=1}^{3}\sum\limits_{j=1}^{3}c_{ij}},
    \label{eq:sds_score}
\end{equation}

\noindent where $c_{ij}$ is the number of elements of class $i$ predicted as the class $j$ and $z$ the number of classes. In particular, $c_{23}$ represents the cells predicted as other when they are elongated and $c_{32}$ are the other cells predicted as elongated.

We used the paired t-test to check the difference between the F1-Score of our results and the state-of-the-art methods. The null hypothesis was that our results were greater. Previously,  the normality of the data distribution was checked, by using the Shapiro-Wilk test.
%In addition to these metrics, we use a Student's t-test. Precisely we use this test to check if we can say significantly than the metrics of our method are greater (and better) than the rest. 
% In addition to these metrics, we use the \textbf{Wilcoxon Test}, that is  a non-parametric test presented by Wilcoxon~\cite{wilcoxon1992individual}. This test is used to determine whether two dependent, and paired, samples were selected from populations having the same underlying distribution. 

\subsection{State-of-the-art Methods}
\label{sec:state_of_art}

In the introduction section we made an study of the methods of the state-of-the-art that separate overlapped objects by finding concave points. To perform our experiments, we selected a representative subset of them.We excluded the methods we could not reproduce due to the absence of information on the original paper and the lack of access to the source code.

We used the original code of  Zafari \emph{et al.}~\cite{zafari2015segmentation} and Gonz{\'a}lez-Hidalgo \emph{et al.}~\cite{gonzalez2014red}. In addition to the two previously named methods, we considered the following methods: 
LaTorre \emph{et al.}~\cite{latorre2013segmentation}, Fern{\'a}ndez  \emph{et al.}~\cite{fernandez1995new}, Song and Wang~\cite{song2009new},
Chaves \emph{et al.}~\cite{chaves2015concave},
Bai \emph{et al.}~\cite{bai2009splitting}, Wang \emph{et al.}~\cite{wang2012clump} and Zafari \emph{et al.}~\cite{zafari2020resolving}. %As we stated before, we implemented these methods using the information available on their respective papers. 

As we did not have the values of the hyperparameters of all of the methods and in order to make a fair comparison, we performed an exhaustive search to obtain the hyperparameters of each method for each experiment, see table \ref{tab:parameters_methods}, even if we had their original values. %, in order to  compare them in a fair way.

%In addition to the state-of-art methods we define a baseline. In this method we assume that a concave point is the point that maximizes the distance between the contour and each convexity defect of the convex hull. The idea is to have a simple method to compare and understand the performance and results of the other proposals.

\begin{table}[!htb]

\small
\centering
\begin{tabular}{cccc}

\hline
Method & \makecell{Original \\ hyperparameters} & \makecell{\miace \\ hyperparameters} &  \makecell{\real \\ hyperparameters} \\ 
\hline
\makecell[t]{Proposed method} & - & \makecell[t]{k: $7$ px, $l_{min}$: $2$ px, \\ $l_{max}$: $11$ px, $\epsilon$: 0.2} &  \makecell[t]{k: $9$ px, $l_{min}$: $6$ px,  \\ $l_{max}$: $25$ px, $\epsilon$: $0.1$} \\  \rule{0pt}{4ex} 
\makecell[t]{LaTorre \emph{et al.}~\cite{latorre2013segmentation}} & \makecell[t]{Min. concavity \\ area: $20$ px, \\ Min. distance \\ to BB: $5$  px, \\ Degree: $10\%$} & \makecell[t]{Min. concavity \\ area: $55$  px, \\ Min. distance \\ to BB: $1$ px, \\ Degree: $10\%$} & \makecell[t]{Min. concavity \\area: $40$  px, \\ Min. distance \\ to BB: $15$  px, \\ Degree: $10\%$} \\ \rule{0pt}{4ex} 

\makecell[t]{Fern{\'a}ndez  \emph{et al.}~\cite{fernandez1995new}} & \makecell[t]{Environment\\ size: $5\times5$, \\ Concavity\\ threshold: -}  & \makecell[t]{Environment\\ size: $7\times7$, \\ Concavity\\ threshold: $1.9$} & \makecell[t]{Environment\\ size: $5\times5$,\\ Concavity\\ threshold: $1$} \\ \rule{0pt}{4ex} 

\makecell[t]{Gonz{\'a}lez-Hidalgo \\ \emph{et al.}~\cite{gonzalez2014red}}  & k: $17$ px, sT:$0.14$ & k: $13$ px, sT: $0.3$ & k: $17px$, sT:$0.14$ \\ \rule{0pt}{4ex} 
\makecell[t]{Zafari \emph{et al.}~\cite{zafari2015segmentation}} & k: $15$ px & k: $5$ px & k: $20$ px   \\ \rule{0pt}{4ex} 
\makecell[t]{Chaves \emph{et al.}~\cite{chaves2015concave}} & \makecell[t]{$\epsilon$: -, k: $2$ px, \\ Concavity\\ threshold: $\pi / 2$} & \makecell[t]{$\epsilon$: 0.1, k: $11$ px, \\ Concavity\\ threshold: $0.$} & \makecell[t]{$\epsilon$: $0.5$, k: $2$ px, \\ Concavity\\ threshold: $0.5$} \\ \rule{0pt}{4ex} 
\makecell[t]{Bai \emph{et al.}~\cite{bai2009splitting}} & \makecell[t]{dTh: $3 $px,\\ Point distance: $1$ px,\\ nStep: $2$ px}    & \makecell[t]{dTh: $1$ px,\\ Point distance: $7$ px,\\ nStep: $3$ px} & \makecell[t]{dTh: $1$ px,\\ Point distance: $2$ px,\\ nStep: $2$ px}                              \\ 

\hline
\end{tabular}
\caption{First column summarize the original parameters of each method. Second and third column summarize the set of hyperparameters for {\sl Experiment 1} and {\sl Experiment 2} respectively.}
\label{tab:parameters_methods}
\end{table}

\renewcommand{\arraystretch}{1} % Default value: 1

\subsection{Experiments}
We developed two experiments to study two different characteristics of the proposed method. First, the precision of the concave points detection. Second, how the precision of the detection affects the posterior segmentation of the overlapped objects. 

%To achieve these objectives we use both \miace dataset (Experiment 1) and \textit{ErythrocytesIDB} dataset (Experiment 2). 
%In order to evaluate the quality of the proposed method and compare it to the state-of-the-art we designed two experiments.  The first one was designed to detect the spatial precision for the concave point prediction algorithms. The second one was designed to evaluate the proposed concave point detector with a well-known application, such as the splitting of overlapped cells in microscopic images of peripheral blood smear samples of red blood cells.

\subsubsection{Experiment 1}

This experiment aimed to compare the detection capacity and the spatial precision of the proposed method with the state-of-the-art. We used the \miace dataset that we generated, because it contains the position of each concave point. Training and test sets were constructed by randomly selecting 1000 images, it is important to notice that intersection between both sets is empty. 

In order to evaluate the performance of each method we used two different performance measures from section \ref{sec:performance_measures}: the Mean Euclidean Distance (MED) and the F1-Score. To compute the F1-Score we matched each detected concave point with a ground truth point. We matched two points if its distance was smaller than an integer threshold, $\theta$,  we set it experimentally. Furthermore, if there were more than one candidate we selected the nearest one. We considered a false positive the predicted points that had not been matched with a ground truth point. We considered a false negative when there was not a candidate for a ground truth point.

To evaluate the performance of each method, we searched the best set of hyperparameters to maximize the F1-score. We performed this process for each $\theta$ between 1 and 20, this allows us to observe the evolution of the performance of each method when we set different thresholds.

\subsubsection{Experiment 2}

This experiment was designed to determine how the precision of concave point detection affect the division of overlapped objects in a real world scenario. We used the ellipse fitting method proposed by Gonz{\`a}lez \emph{et al.} \cite{gonzalez2014red} to divide the overlapped objects from the detected concave points. After we completed this step we compared the ground truth with the predicted objects. In this experiment we used the \textit{ErythrocytesIDB2} dataset. As a training set we randomly selected the $70$\% of the images, there are 34 images that contains 1825 cells. The remaining $30$\% of the images, 16 images containing 980 cells, were used as test set.

\begin{figure*}[!hptb]
\centering
 \begin{subfigure}[b]{0.23\linewidth}
    \centering
    \includegraphics[width=0.99\textwidth]{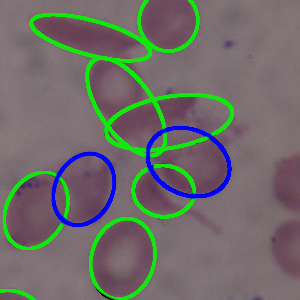}
 \end{subfigure}%   
 \hspace{0.15cm}
 \begin{subfigure}[b]{0.23\linewidth}
    \centering
    \includegraphics[width=0.99\textwidth]{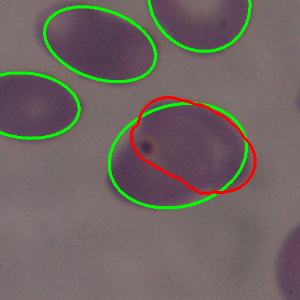}
 \end{subfigure}%
 \hspace{0.15cm}
 \begin{subfigure}[b]{0.23\linewidth}
    \centering
    \includegraphics[width=0.99\textwidth]{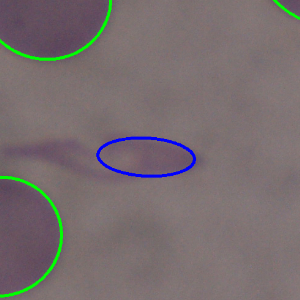}
 \end{subfigure}%
 \hspace{0.15cm}
 \begin{subfigure}[b]{0.23\linewidth}
    \centering
    \includegraphics[width=0.99\textwidth]{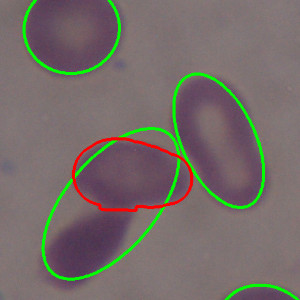}
 \end{subfigure}%
  \caption{Examples of false positives and false negatives using the proposed algorithm. Detection of none existing cells in blue color. Cells from the ground truth that are not detected in red color. Correct segmentation in green color.}\label{fig:ghost_cell}
\end{figure*}

The problem addressed on this experiment was a multi-class problem, for this reason we used the CBA, MCC, and SDS-Score and the adapted version of the F1-Score averaging the results for each class. We considered the prediction of a non existing cell in the ground truth as false positive, and the omission to predict an existing cell in the ground truth as a false negative. Figure \ref{fig:ghost_cell} depicts some examples of these false positives and false negatives detections.
% Furthermore, we also provide the proportion between these two cases and the total amount of detected cells.

%Another possibility that the F1-Score has not in consideration are the prediction of elements that not exists and the omission to predict some objects that it exists. In our case this is possible as can be seen in figure \ref{fig:ghost_cell}. These two special cases are not considered for the calculation of the metrics, instead, we present the ratio between the occurrence of each type of error and the normal cases. We called these situations \textit{over sample} and \textit{under-sample}, respectively. Furthermore, to facilitate the study of the results we aggregate both ratios with the harmonic mean. 

\section{Results and discussion} \label{results}

In this section we analyze and discuss the results of the experiments over the datasets described in section \ref{sec:datasets}.
%We developed two experiments to study two different characteristics of the proposed method. First, the precision of the concave points detection. Second, how the precision of the detection affects the posterior segmentation of the overlapped objects. %To achieve these objectives we use both \miace dataset (Experiment 1) and \textit{ErythrocytesIDB} dataset (Experiment 2). 

%As we pointed previously, see Section \ref{sec:state_of_art}, in order to make a fair comparison of the multiple proposals, we fine tuned the hyperparameters of each method for each experiment, its values are summarized on Table \ref{tab:parameters_methods}.

%\ref{tab:train-results-miace} and \ref{tab:results-f1score-train}.  \ref{tab:sample_errors_table}. These tables contain the metrics already described and some extra information as the Precision and Recall, metrics used to build the F1-Score.

 %Both tables indicate the results of the \miace dataset. We separate it into two table for train and test sets. In both tables 
 
\subsection{Experiment 1}

\begin{figure*}[!htb]
\begin{center}
	\includegraphics[width = \linewidth]{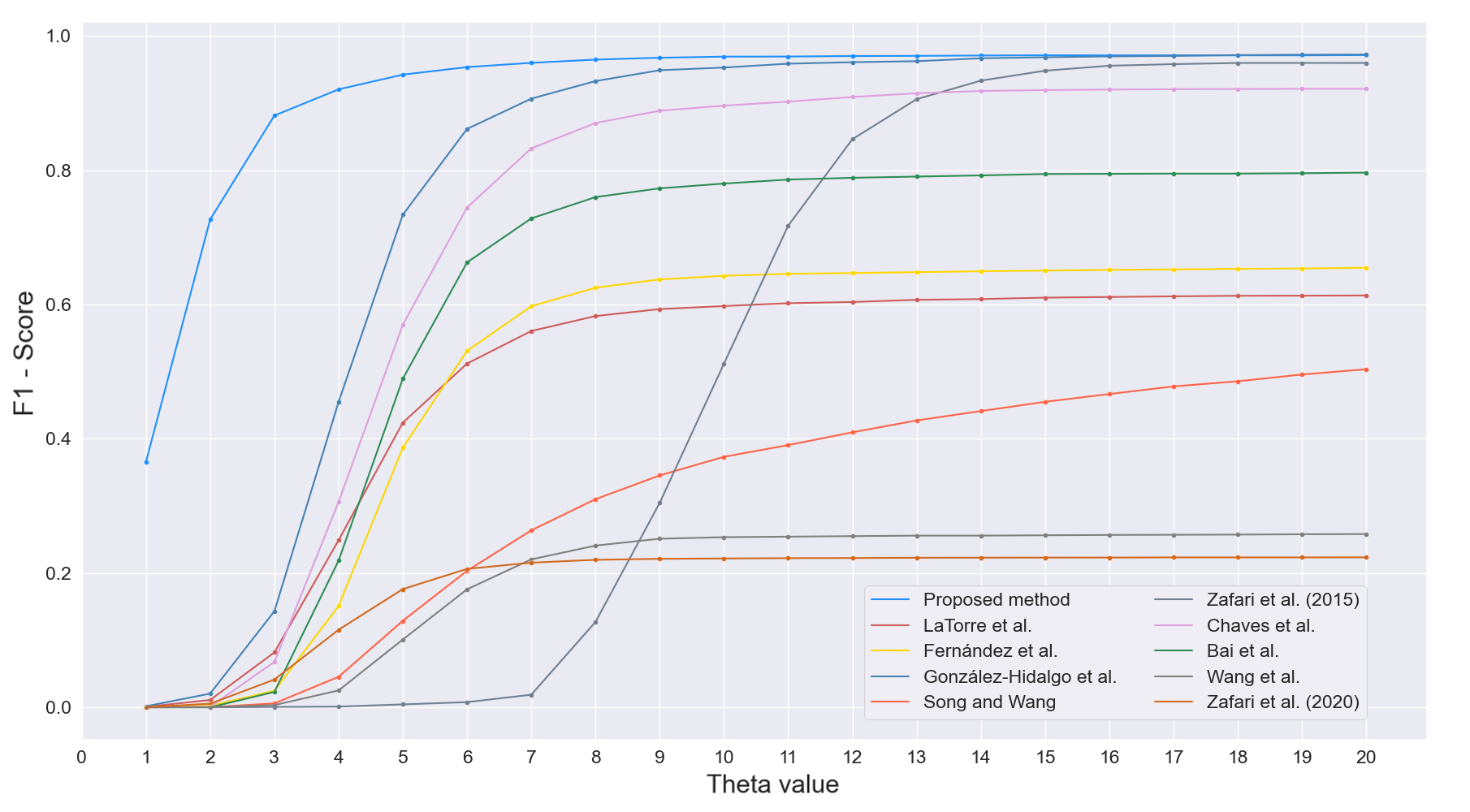}
\end{center}
\caption{F1-score on the test set of \miace dataset with different values of $\theta$,  that is the maximum allowed distance to make a match between a detection and a ground-truth point. The proposed method outperforms the methods described in section 4.3 for all $\theta$ values.}\label{fig:thetas}
\end{figure*}

Figure \ref{fig:thetas} depicts the F1-score we obtained with the test set for each value of $\theta$ and the best hyperparameters for each method. We can observe that the proposed method outperforms the results of the state-of-art, the difference is bigger when $\theta$ is lower than 10 pixels, when is more difficult to make a match between a detection and a ground-truth point.

Tables \ref{tab:train-results-miace} and \ref{tab:results-miace} outline the results we obtained for the detection of concave points in {\sl Experiment 1} when $\theta = 15px $. We selected this value according to results depicted in figure \ref{fig:thetas} where all methods have an stable F1-score. The tables summarize the precision, recall, F1-Score and $MED$ values of the concave point detection methods. We also added the standard deviation ($STD$) of the $MED$ measure in order to provide complementary information. In our evaluation it is important to obtain an small value in $MED$ but it was also important to ensure that this measure was not scattered.

Table \ref{tab:train-results-miace} summarize the results obtained for the training set. We can observe that the proposed method achieved the best value for the F1-Score and it is almost tied with Gonz{\'a}lez-Hidalgo \emph{et al.} method. The methods of Chaves~\emph{et al.} and Zafari~\emph{et al.}~\cite{zafari2015segmentation} also had good results for this measure. The other methods had an strong unbalance between results from the precision and recall, higher precision usually provokes a lower recall value and vice-versa. When this situation occur the methods obtained a low F1-Score performance. Regarding to the $MED$ measure, the proposed method obtained the \textst{second} best result with the lowest standard deviation. This indicates that the values are close to the mean and less scattered than the others.

\begin{table*}[!hbt]
\centering
\begin{tabular}{lccccc}
\hline
Method                                                      & Precision        & Recall            & F1-Score          & MED               & STD \\ \hline
\textbf{Proposed method}                                    & \textbf{0.998}   & 0.943             & \textbf{0.969}    & \textbf{1.777}             & \textbf{1.551}      \\
LaTorre \emph{et al.}~\cite{latorre2013segmentation}        & 0.445            & 0.928             & 0.604             & 9.270             & 20.226      \\
Fern{\'a}ndez  \emph{et al.}~\cite{fernandez1995new}        & 0.601            & 0.701             & 0.646             & 23.823            & 41.759      \\
Gonz{\'a}lez-Hidalgo \emph{et al.}~\cite{gonzalez2014red}   & 0.982            & 0.953             & 0.967             & 4.544             & 3.682       \\
Song and Wang~\cite{song2009new}                            & 0.481            & 0.421             & 0.449             & 41.213            & 51.764      \\
Zafari \emph{et al.}~\cite{zafari2015segmentation}          & 0.978            & 0.911             & 0.943             & 10.193            & 4.337       \\
Chaves \emph{et al.}~\cite{chaves2015concave}               & 0.991            & 0.847             & 0.913             & 5.404             & 5.795      \\
Bai \emph{et al.}~\cite{bai2009splitting}                   & 0.992            & 0.646             & 0.783             & 5.0948            & 3.057      \\
Wang \emph{et al.}~\cite{wang2012clump}                     & 0.275            & 0.255             & 0.264             & 83.742            & 62.325      \\ 
Zafari \emph{et al.}~\cite{zafari2020resolving}             & 0.128            & \textbf{0.984}    & 0.226             & 4.470             & 3.120      \\ 
\hline
\end{tabular}
\caption{Results of the {\sl Experiment 1} using 1000 images of the training set from synthetic dataset. \textit{MED} is the mean of the euclidean distance from a detected point to the closest ground truth point, \textit{STD} is its standard deviation. %In bold, the best result of each metric.
}
\label{tab:train-results-miace}
\end{table*}

Table \ref{tab:results-miace} sum up the results obtained for the test set. We can observe that the results were very similar than the ones we obtained using the training set. Also, the proposed method achieved the best F1-Score and $MED$ with a very low $STD$.  %s are very different, 3.152 for the proposed method versus 10.206 for the second one.

\begin{table*}[!hbt]
\centering
\begin{tabular}{lccccc}
\hline
Method                                                      & Precision        & Recall            & F1-Score          & MED               & STD \\ \hline
\textbf{Proposed method}                                    & \textbf{0.995}   & 0.948             & \textbf{0.971}    & \textbf{1.762}             & \textbf{2.023}        \\
LaTorre \emph{et al.}~\cite{latorre2013segmentation}        & 0.454            & 0.931             & 0.610             & 9.129             & 20.223      \\
Fern{\'a}ndez  \emph{et al.}~\cite{fernandez1995new}        & 0.597            & 0.714             & 0.650             & 22.305            & 39.853      \\
Gonz{\'a}lez-Hidalgo \emph{et al.}~\cite{gonzalez2014red}   & 0.977            & 0.959             & 0.968             & 4.504             & 3.002       \\
Song and Wang~\cite{song2009new}                            & 0.511            & 0.461             & 0.484             & 38.156            & 51.613      \\
Zafari \emph{et al.}~\cite{zafari2015segmentation}          & 0.979            & 0.919             & 0.948             & 10.176            & 5.326       \\
Chaves \emph{et al.}~\cite{chaves2015concave}               & 0.987            & 0.859             & 0.919             & 5.414             & 5.707       \\
Bai \emph{et al.}~\cite{bai2009splitting}                   & 0.989            & 0.664             & 0.794             & 5.338             & 6.276       \\
Wang \emph{et al.}~\cite{wang2012clump}                     & 0.239            & 0.221             & 0.229             & 96.446            & 70.646      \\ 
Zafari \emph{et al.}~\cite{zafari2020resolving}             & 0.125            & \textbf{0.986}    & 0.223             & 4.462             & 3.443       \\ 
\hline
\end{tabular}
\caption{Results of the {\sl Experiment 1} using 1000 images of the test set from synthetic dataset. \textit{MED} is the mean of the euclidean distance from a detected point to the closest ground truth point, \textit{STD} is its standard deviation. %In bold, the best result of each metric.
}
\label{tab:results-miace}
\end{table*}

From the previous analysis we can state that the proposed method find the concave points with the highest balance between the precision and the recall, that means lower rates of false positives and false negatives. The lower values on $MED$ and $STD$ metrics denote that the detected points are close to the ground truth ones, that means a high degree of spatial precision.
 
\subsection{Experiment 2}
We summarized the results for {\sl Experiment 2} in tables \ref{tab:results-f1score-train},  \ref{tab:results-f1score} obtained with the images from \real dataset.  As in  the previous experiment, we separated the results in two different tables, for the training set and the test set, respectively. 
The results for the training set are in Table \ref{tab:results-f1score-train}.  We can see that the proposed method was surpassed in the CBA measure by the method described in Bai~\emph{et al.}
but had the best results in all the other metrics. Table \ref{tab:results-f1score} shows the results obtained for the test set. It should be noted that when the proposed method had been faced with unseen data it  achieved the best values in all metrics. %and increases the distance with the rest of the proposals  \textcolor{red}{respecting} the results obtained using the training set. 

\begin{figure}
\begin{tabular}{>{\centering\arraybackslash}p{0.3\linewidth} >{\centering\arraybackslash}p{0.3\linewidth}>{\centering\arraybackslash}p{0.3\linewidth}} 
  \includegraphics[width=0.85\linewidth]{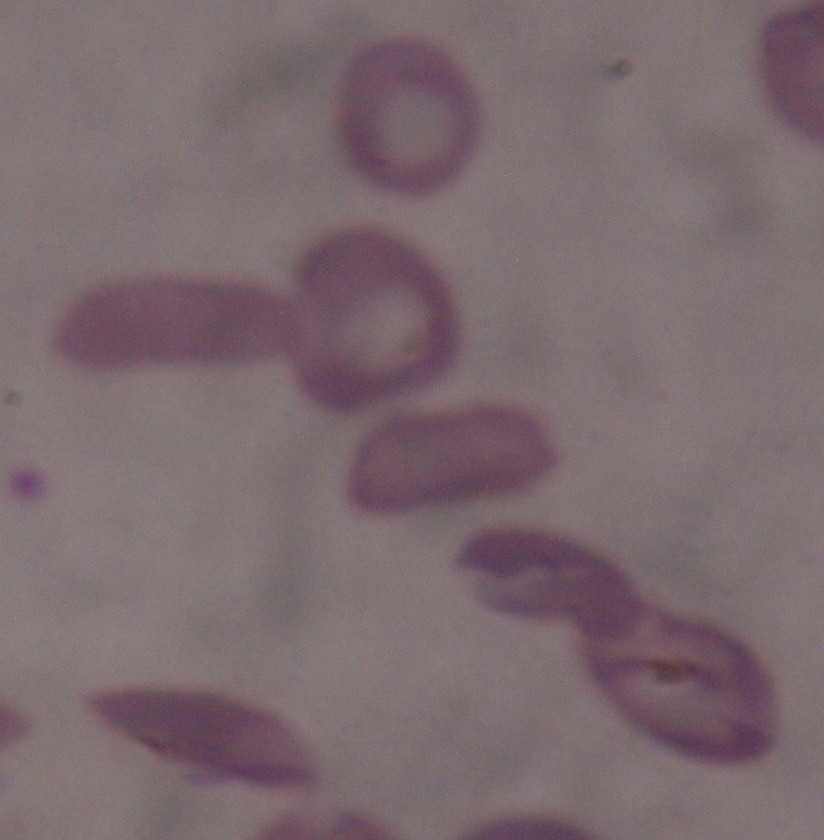} &
  \includegraphics[width=0.85\linewidth]{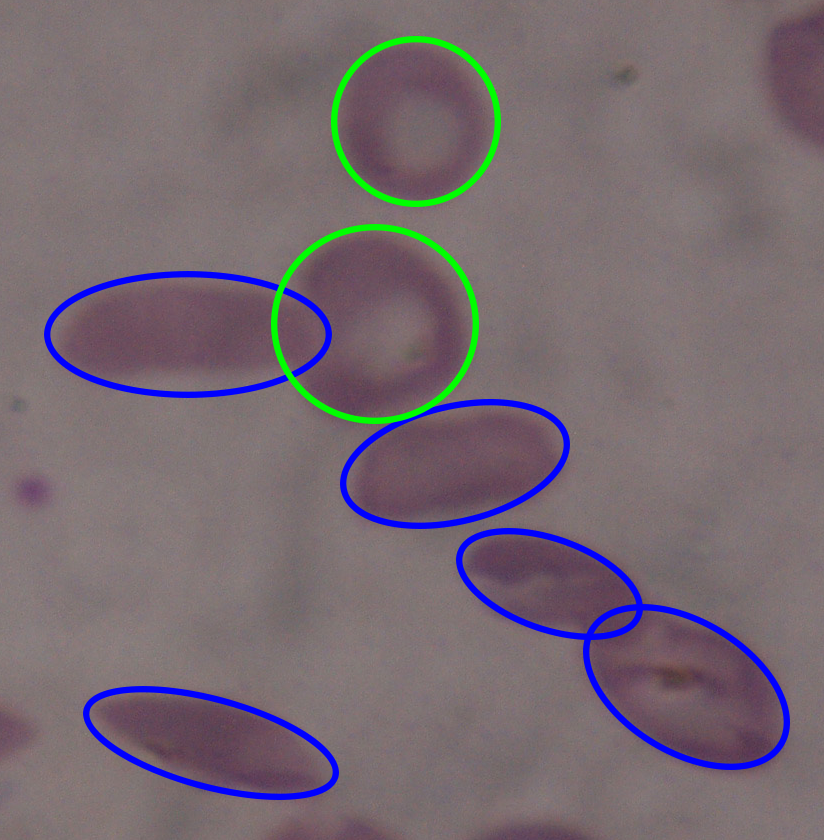}  &   \includegraphics[width=0.85\linewidth]{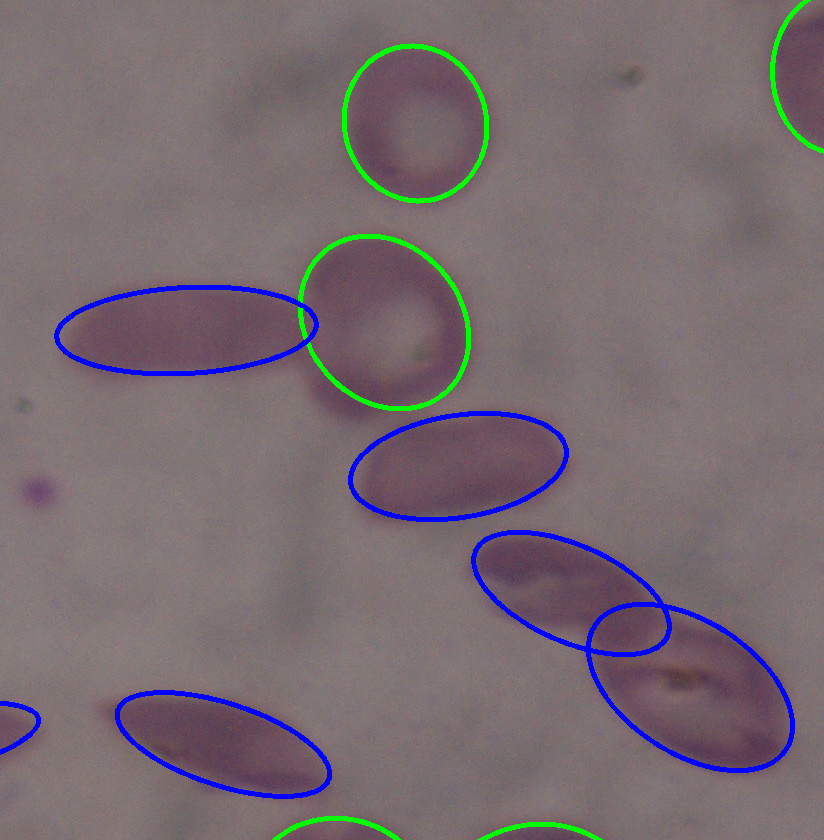}  \\ 
  Original image & Ground truth & Proposed method \\[6pt]
 \includegraphics[width=0.85\linewidth]{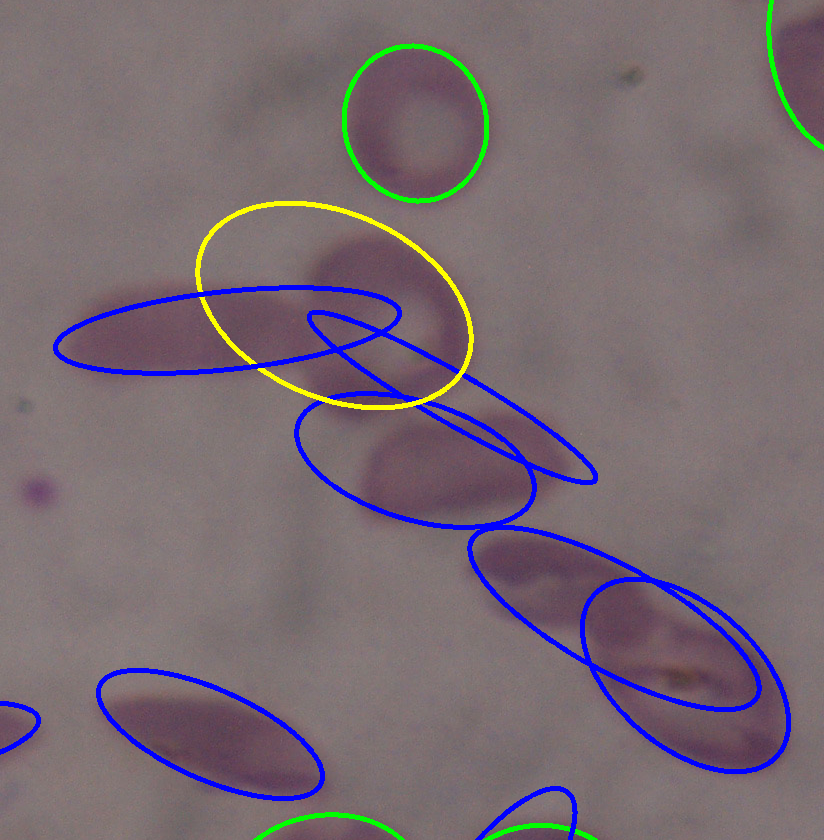}    &  \includegraphics[width=0.85\linewidth]{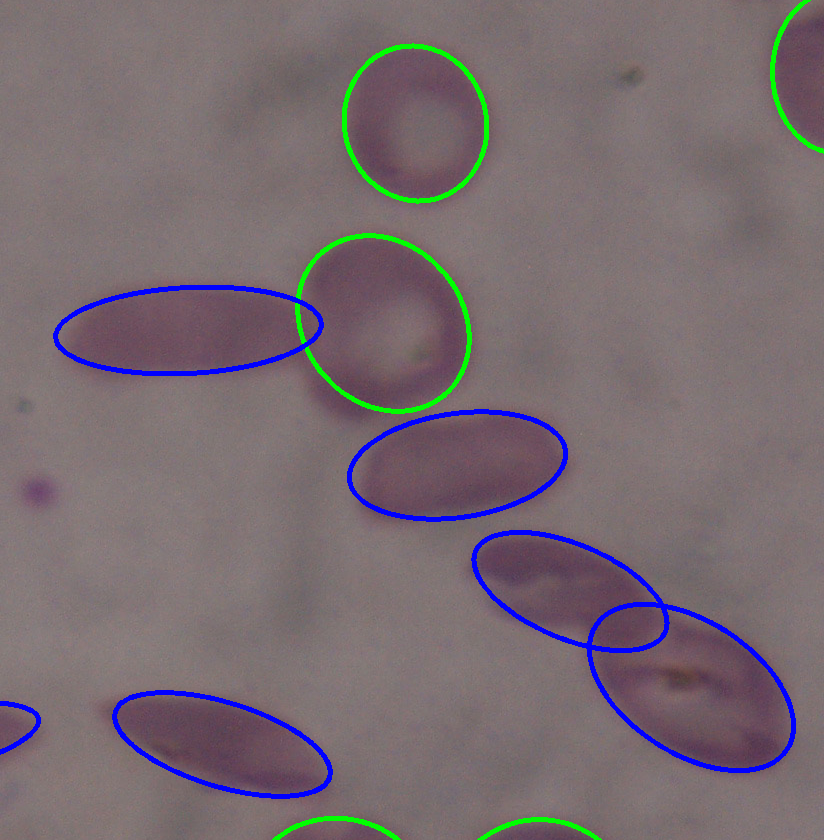} & \includegraphics[width=0.85\linewidth]{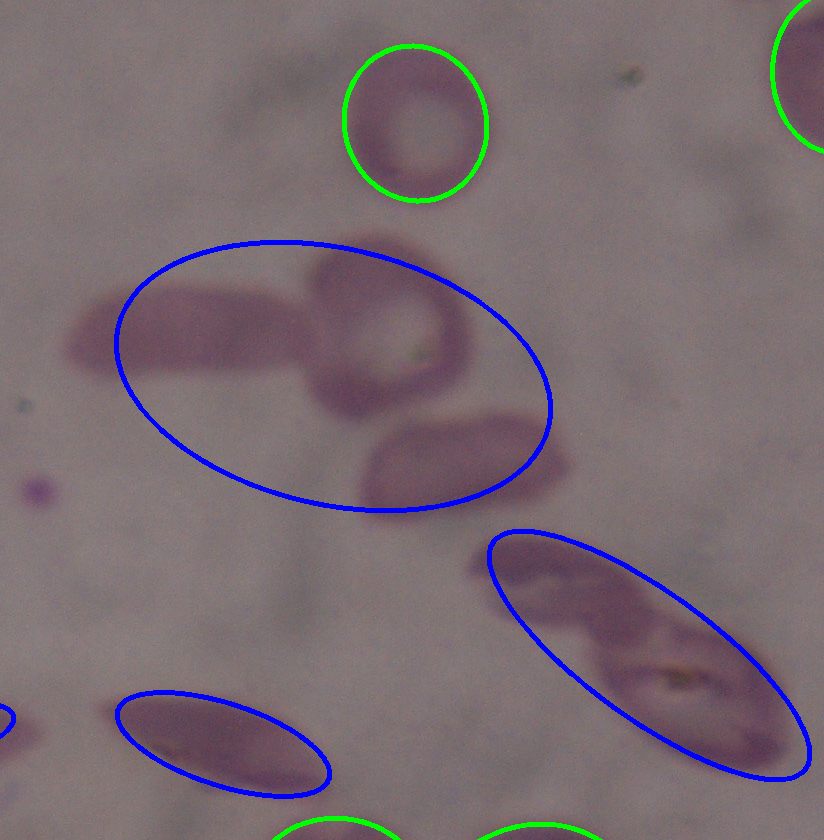}\\
  Fern{\'a}ndez~\emph{et al.}~\cite{fernandez1995new}   & Gonz{\'a}lez-Hidalgo \emph{et al.}~\cite{gonzalez2014red} & Song and Wang~\cite{song2009new} \\[6pt]
  \includegraphics[width=0.85\linewidth]{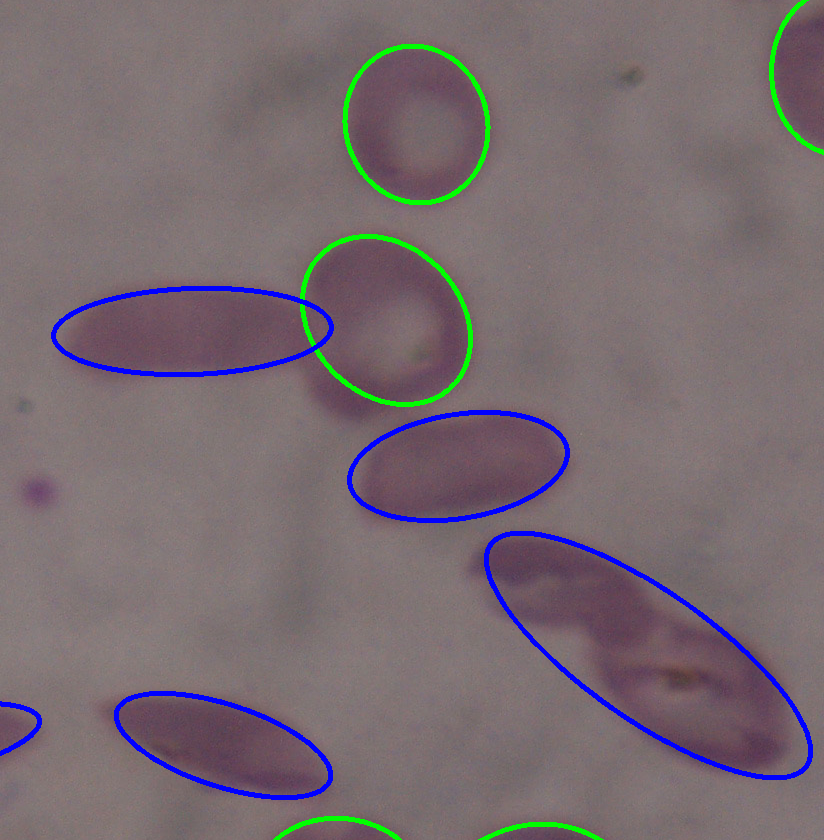}    &  \includegraphics[width=0.85\linewidth]{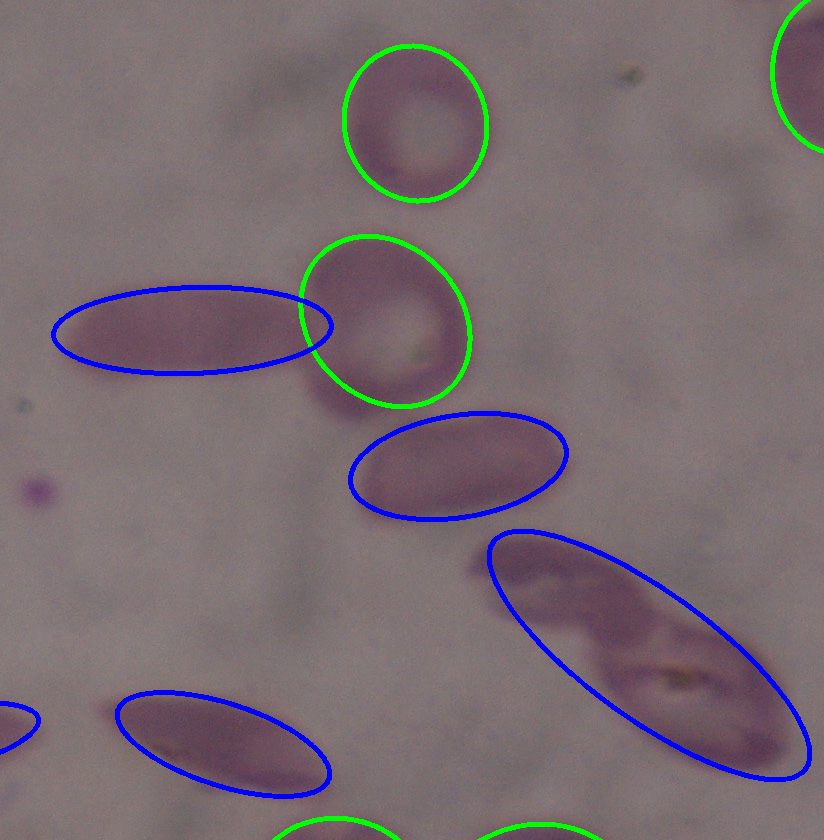} & \includegraphics[width=0.85\linewidth]{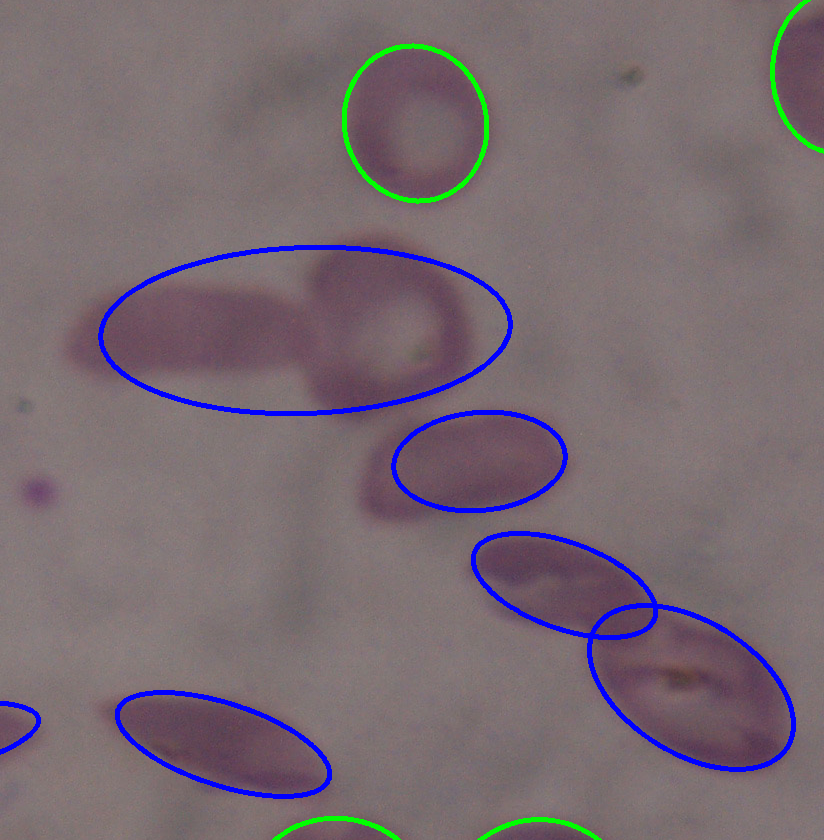}\\
  Zafari~\emph{et al.}~\cite{zafari2015segmentation}  & Chaves~\emph{et al.}~\cite{chaves2015concave} & Bai~\emph{et al.}~\cite{bai2009splitting}  \\[6pt]
  \includegraphics[width=0.85\linewidth]{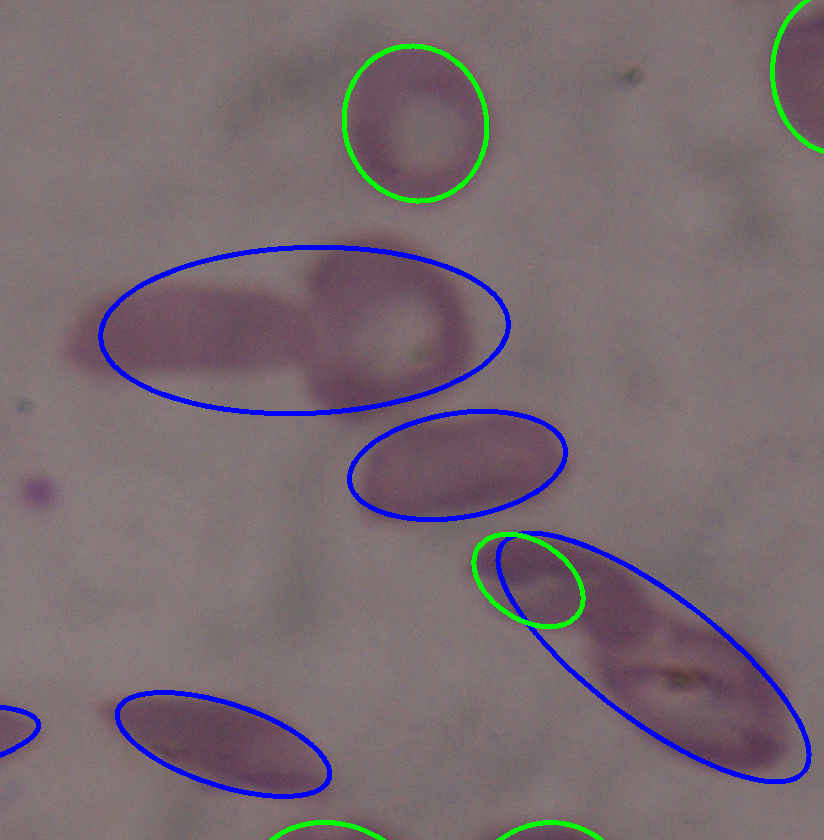}     &  \includegraphics[width=0.85\linewidth]{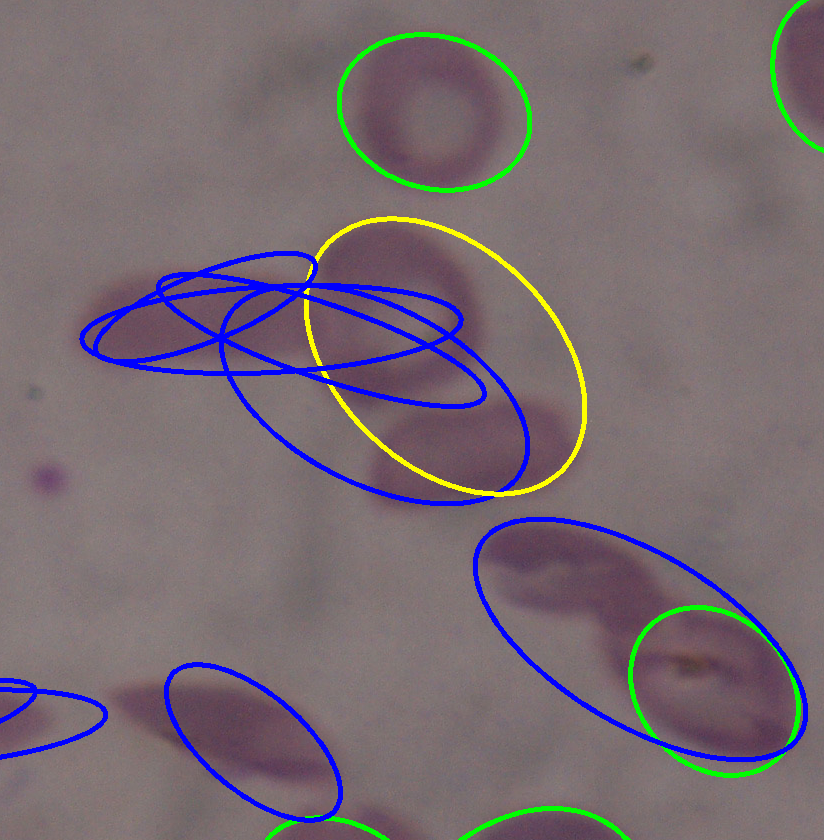}   & 
  \includegraphics[width=0.85\linewidth]{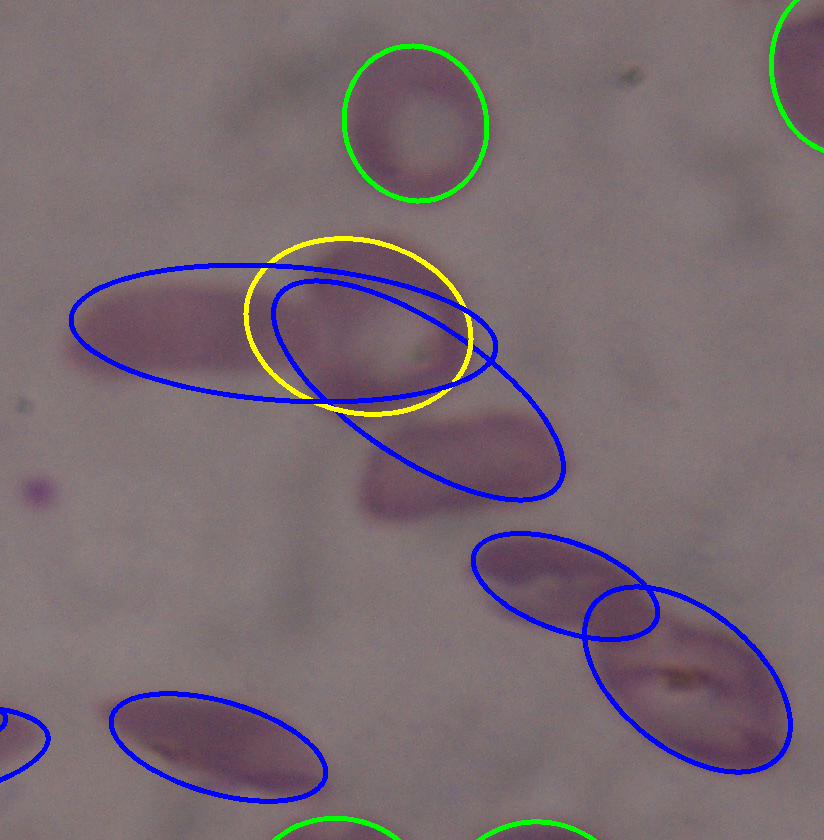} \\
  Wang~\emph{et al.}~\cite{wang2012clump} & Zafari~\emph{et al.}~\cite{zafari2020resolving}  &  LaTorre \emph{et al.}~\cite{latorre2013segmentation} \\[6pt]
% \multicolumn{3}{>{\centering\arraybackslash}  p{1\linewidth}}{\includegraphics[width=0.25\linewidth]{figures/cmp/B.png} }\\
% \multicolumn{3}{>{\centering\arraybackslash} p{1\linewidth}}{ Wang~\emph{et al.}~\cite{wang2012clump} }
\end{tabular}
\caption{Example of results obtained from the test set of \real dataset for each method. Colors used in this figure: blue for the elongated cells, green for the circular cells, yellow for the other cells. %The proposed method and the one by Gonz{\'a}lez-Hidalgo \emph{et al.} are the
Only two methods can segment both clusters correctly.}

\label{fig:grid_imgs}
\end{figure}

Figure \ref{fig:grid_imgs} depicts some results of the {\sl Experiment 2} for an image in the test set. In the original image we can observe that there are two different clusters of cells. The proposed method and the one by Gonz{\'a}lez-Hidalgo \emph{et al.} were the only ones capable to segment both clusters correctly, that shows the difficulty to solve this problem as we had the overlapping zones that did not provide any information. In that figure we can observe that the application of different concave point detection methods led to different object segmentation results, it is necessary to point out that we used the same algorithm to segment the clusters for all the approaches.
 %\textcolor{red}{We can conclude that the proposed method is a more general method, that can be applied in unseen data without losing quality.} 

% !!!!!!!!!!!!!!!!!!!!!!! Imatge 6 !!!!!!!!!!!!!!!!!!!!!!!!!!!!

We used the paired t-test to compare the methods and to determine if the proposed one outperforms the other ones significantly. %since it is a  statistical hypothesis test after testing the normality hypothesis of the data. 
The numerical results of the comparison of the algorithms for the F1-score performance measure are summarised in Table \ref{tab:ttest}. Using the F1-score, we compared our method with each one of the other methods using the t-test with a confidence level of $95\%$. The result of the application of this test is that the proposed method outperformed significantly most of the other proposals, except the ones based on curvature estimation as Gonz{\'a}lez-Hidalgo~\emph{et al.} \cite{gonzalez2014red}, Chaves~\emph{et al.} \cite{chaves2015concave} and Bai~\emph{et al.}  \cite{bai2009splitting}, where the improvement was not statistically significant.

% Table \ref{tab:sample_errors_table} depicts two possible errors of each method, the detection of non existing cells and cells that should been detected but it hasn't. This values are indicated as a proportion. The table also contains the number of correct predictions. This s

From the analysis of the results obtained with {\sl Experiment 2}, we can determine that the increase on the precision in the detection of the concave points implies an improvement in the results in the segmentation of the overlapping objects.

\iffalse
From the obtained results  on the second experiment, we can conclude that  the increase on the precision of the concave point detection improves the result of a later segmentation method. %Our proposed method surpasses the rest of the method in the almost all used metrics.
\fi

\begin{table*}[!hbt]
\centering
\begin{tabular}{lcccccc}
\hline
Method                                                  & Precision      & Recall           & F1-Score       & SDS-Score        & CBA           & MCC \\ \hline
\textbf{Proposed method}                                & 0.862         & \textbf{0.882}   & \textbf{0.872} & \textbf{0.908}   & 0.659         & \textbf{0.739}  \\
LaTorre \emph{et al.}~\cite{latorre2013segmentation}    & 0.813          & 0.806            & 0.808          & 0.873            & 0.637         & 0.642 \\
Fern{\'a}ndez \emph{et al.}~\cite{fernandez1995new}     & 0.777          & 0.789            & 0.781          & 0.846            & 0.569         & 0.587 \\
Gonz{\'a}lez-Hidalgo \emph{et al.}~\cite{gonzalez2014red} & 0.847        & 0.852            & 0.850          & 0.902            & 0.649          & 0.703 \\
%Baseline                                                & 0.844          & 0.847            & 0.845          & 0.891            & 0.633          & 0.682 \\
Song and Wang~\cite{song2009new}                        & 0.669          & 0.849            & 0.746          & 0.842            & 0.545          & 0.595 \\
Zafari \emph{et al.}~\cite{zafari2015segmentation}      & 0.787          & 0.871            & 0.826          & 0.881            & 0.636          & 0.687  \\
Chaves \emph{et al.}~\cite{chaves2015concave}           & 0.848          & 0.873            & 0.860          & 0.901            & 0.648          & 0.723 \\
Bai \emph{et al.}~\cite{bai2009splitting}               & \textbf{0.871} & 0.870            & 0.870          & 0.907            & \textbf{0.667} & 0.731 \\
Wang \emph{et al.}~\cite{wang2012clump}                 & 0.729          & 0.842            & 0.781          & 0.856            & 0.594          & 0.621 \\
Zafari \emph{et al.}~\cite{zafari2020resolving}         & 0.704          & 0.783            & 0.724          & 0.822            & 0.509          & 0.563 \\
\hline
\end{tabular}
\caption{Results of the {\sl Experiment 2} using the 34 images of the train set from \textit{ErythrocytesIDB2}. \textit{SDS} is the sickle cell diagnosis support score. \textit{MCC} is the Matthew's Correlation Coefficient and \textit{CBA} is the Class Balance Accuracy.% In bold, the best result of each metric.
}
\label{tab:results-f1score-train}
\end{table*}

\begin{table*}[!hbt]
\centering
\begin{tabular}{lcccccc}
\hline
Method & Precision & Recall & F1-Score & SDS-Score & CBA & MCC \\ \hline
\textbf{Proposed method} & \textbf{0.856} & \textbf{0.861}   & \textbf{0.858} & \textbf{0.888}   & \textbf{0.665} & \textbf{0.724}   \\
LaTorre \emph{et al.}~\cite{latorre2013segmentation}    & 0.813          & 0.783            & 0.798          & 0.865            & 0.602          & 0.631            \\
Fern{\'a}ndez \emph{et al.}~\cite{fernandez1995new}     & 0.766          & 0.752            & 0.758          & 0.843            & 0.598          & 0.601            \\
Gonz{\'a}lez-Hidalgo \emph{et al.}~\cite{gonzalez2014red}       & 0.838          & 0.821            & 0.829          & 0.877            & 0.625          & 0.677            \\
Song and Wang~\cite{song2009new}                        & 0.665          & 0.825            & 0.735          & 0.829            & 0.592          & 0.595            \\
Zafari \emph{et al.}~\cite{zafari2015segmentation}      & 0.766          & 0.842            & 0.801          & 0.845            & 0.650          &  0.655           \\
Chaves \emph{et al.}~\cite{chaves2015concave}           & 0.826          & 0.846            & 0.836          & 0.873            & 0.634          & 0.684            \\
Bai \emph{et al.}~\cite{bai2009splitting}               & 0.842          & 0.836            & 0.838          & 0.879            & 0.638          & 0.693            \\
Wang \emph{et al.}~\cite{wang2012clump}                 & 0.711          & 0.821            & 0.762          & 0.829            & 0.618          & 0.592            \\
Zafari \emph{et al.}~\cite{zafari2020resolving}         & 0.696          & 0.728            & 0.697          & 0.806            & 0.532          & 0.536            \\
\hline
\end{tabular}
\caption{Results of the {\sl Experiment 2} using the 16 images of the test set from \textit{ErythrocytesIDB2}. \textit{SDS} is the sickle cell diagnosis support score. \textit{MCC} is the Matthew's Correlation Coefficient and \textit{CBA} is the Class Balance Accuracy.% In bold, the best result of each metric.
}
\label{tab:results-f1score}
\end{table*}

\begin{table}[!hbt]
\centering
\begin{tabular}{lcc}
\hline
Methods  & p-value  & t stadistic   \\ 
\hline
LaTorre \emph{et al.}           & $\boldsymbol{2.46 \times10^{-3}}$     & 3.051         \\
Fernández \emph{et al.}         & $\boldsymbol{3.62 \times10^{-5}}$     & 4.682         \\
Gonzàlez-Hidalgo \emph{et al.}  & $\boldsymbol{6.41 \times10^{-2}}$     & 1.565         \\
Song and Wang                   & $\boldsymbol{1.11\times10^{-4}}$      & 4.378         \\
Zafari \emph{et al.}            & \textbf{0.01546}                      & 2.297         \\
Chaves \emph{et al.}            & 0.1693                                & 0.973         \\
Bai \emph{et al.}               & 0.1824                                & 0.922         \\
Wang \emph{et al.}              & $\boldsymbol{5.42\times10^{-4}}$      & 3.707         \\ 
Zafari \emph{et al.}            & $\boldsymbol{7.15\times10^{-7}}$      & 5.516         \\
\hline
\end{tabular}
\caption{Results of applying the t-test between our proposed method and the rest of state-of-the-art methods. The alternative hypothesis is: the mean of our method is greater than the others. %In bold the p-values lower than 0.05.
}

%\caption{Results of applying the t-test between our proposed method F1-Score of the test set from \real dataset and the rest of state-of-the-art methods. The alternative hypothesis is: the mean of our method is greater than the others. %In bold the p-values lower than 0.05.}

%, the threshold that allows to assure with a 95\% of confidence that our proposed method has higher values than the rest.}
\label{tab:ttest}
\end{table}

Finally, we want to remark that the proposed method can be considered transparent~\cite{arrieta2020explainable}, because has the ability of simulatability (being simulated or thought about strictly by a human), decomposability (explaining each of the parts of the method), and algorithmic transparency (the user can understand the process followed by the method to produce any given output from its input data). This is especially important in health, to trust the behavior of intelligent systems.

\section{Conclusions}  \label{conclusion}
%introducció

The concave point detection is a first step to segment overlapped objects in images, the existence of this clusters reduces the information available in some areas of the image, maintaining it an still challenging problem. 

%CANVIAR
The methodology we proposed in this paper is based on the curvature approximation on each point of the contour of the overlapped objects. First, we selected the regions with higher curvature levels because these are the regions with that contains at least one interest point. Second, we applied a recursive algorithm to refine the previous selected regions. Finally, we obtained a concave point from each region.
%First, we identified the concave points as the local extreme of the curvature. As a second step, we selected regions with the highest probability to contain an interest point, that is, regions with higher curvature. Finally we obtained an interest point from each region by applying a recursive algorithm and we selected the concave ones. 

We included as results the implemented code,  the confusion matrices with the raw data of all the studied methods and the images used as training and test sets to allow researchers to more easily compute other metrics, see \url{https://github.com/expainingAI/overlapped-objects}. As an additional contribution we constructed and opened to the scientific community a synthetic dataset to simulate overlapping objects, we provided the position of the concave points as a ground-truth, as far as we know, is the first public dataset containing overlapping objects with annotated concave points.. We used this dataset to compare the capacity of detection and the spatial precision of the proposed method with the state-of-the-art, see \url{https://github.com/expainingAI/overArt}. For the sake of scientific progress, it would be beneficial if authors published their raw data, code and the image datasets that they used.

Finally, as a case study, we evaluated the proposed concave point detector and the state-of-the-art methods with a well-known application, such as the splitting of overlapped cells in microscopic images of peripheral blood smear samples of RBC of patients with sickle-cell disease. The goal of the case study was two check if the spacial precision of the concave points detector method affected the results of a classification algorithm of the morphology of RBC in a real world scenario.

Results from experimentation proved that the proposed method had better results in both synthetic and real datasets, in %to ensure its quality we used 
multiple standard metrics.  We can conclude that the proposed methodology detects concave points with the highest exactitude as it obtain lower values in the \textit{MED} metric and a small standard deviation. Regarding the same experiment it have obtained the best F1-Score, that means a good balance between its precision and recall detecting concave points.  We designed a second experiment to determine how the precision of concave point detection affect the division of overlapped objects in a real world scenario. From its results we can conclude that a method with higher precision to find concave points, as the proposed one in this paper, help to achieve a better cell classification. Finally, it is important to notice this method is not limited to the case study we performed in this paper, it can be used for other applications where separation between overlapping objects is required. %  helps to achieve the best results in all metrics for this experiment.}
Furthermore, the methods based on the detection of concave points can perform a good
segmentation without using a big dataset and without input image size constraints unlike
deep learning methods. Moreover, the detection of concave points for segmenting overlapped
objects can be considered to be transparent because presents simulatability, decomposability
and algorithmic transparency \cite{arrieta2020explainable}.

\section*{Author Statement}
\noindent \textbf{Miquel Miró-Nicolau}: Software, Visualization, Formal analysis , Writing - Original Draft. \textbf{ Gabriel Moyà Alcover}: Conceptualization, Validation, Project administration. \textbf{ Manuel González-Hidalgo}: Methodology , Formal analysis, Writing - Review  \& Editing. \textbf{Antoni Jaume-i-Capó}: Methodology, Writing - Review \& Editing, Resources.

\section*{Acknowledgments}
This work was funded by project EXPLainable Artificial INtelligence systems for health and well-beING (EXPLAINING) (PID2019-104829RA-I00 / AEI / 10.13039/501100011033), the Spanish Grant FEDER/Ministerio de Econom\'ia, Industria y Competitividad - AEI/TIN2016-75404-P. Miquel Miró also benefited from the fellowship FPI/035/2020 (Govern de les Illes Balears) %under an operational program co-financed by the European Social Fund.
\bibliographystyle{spmpsci} 
\bibliography{bibliografia} %Actualiza para tu caso

\begin{thebibliography}{10}
\providecommand{\url}[1]{{#1}}
\providecommand{\urlprefix}{URL }
\expandafter\ifx\csname urlstyle\endcsname\relax
  \providecommand{\doi}[1]{DOI~\discretionary{}{}{}#1}\else
  \providecommand{\doi}{DOI~\discretionary{}{}{}\begingroup
  \urlstyle{rm}\Url}\fi

\bibitem{arrieta2020explainable}
Arrieta, A.B., D{\'\i}az-Rodr{\'\i}guez, N., Del~Ser, J., Bennetot, A., Tabik,
  S., Barbado, A., Garc{\'\i}a, S., Gil-L{\'o}pez, S., Molina, D., Benjamins,
  R., et~al.: Explainable artificial intelligence (xai): Concepts, taxonomies,
  opportunities and challenges toward responsible ai.
\newblock Information Fusion \textbf{58}, 82--115 (2020)

\bibitem{bai2009splitting}
Bai, X., Sun, C., Zhou, F.: Splitting touching cells based on concave points
  and ellipse fitting.
\newblock Pattern recognition \textbf{42}(11), 2434--2446 (2009)

\bibitem{chan1999active}
Chan, T., Vese, L.: An active contour model without edges.
\newblock In: International Conference on Scale-Space Theories in Computer
  Vision, pp. 141--151. Springer (1999)

\bibitem{chang2007segmentation}
Chang, H., Yang, Q., Parvin, B.: Segmentation of heterogeneous blob objects
  through voting and level set formulation.
\newblock Pattern recognition letters \textbf{28}(13), 1781--1787 (2007)

\bibitem{chaves2015concave}
Chaves, D., Trujillo, M., Barraza, J.M.: Concave points for separating touching
  particles.
\newblock In: Sixth International Conference on Graphic and Image Processing
  (ICGIP 2014), vol. 9443, p. 94431W. International Society for Optics and
  Photonics (2015)

\bibitem{delgado2020diagnosis}
Delgado-Font, W., Escobedo-Nicot, M., Gonz{\'a}lez-Hidalgo, M., Herold-Garcia,
  S., Jaume-i Capo, A., Mir, A.: Diagnosis support of sickle cell anemia by
  classifying red blood cell shape in peripheral blood images.
\newblock Medical \& Biological Engineering \& Computing  (2020)

\bibitem{douglas1973algorithms}
Douglas, D.H., Peucker, T.K.: Algorithms for the reduction of the number of
  points required to represent a digitized line or its caricature.
\newblock Cartographica: the international journal for geographic information
  and geovisualization \textbf{10}(2), 112--122 (1973)

\bibitem{farhan2013novel}
Farhan, M., Yli-Harja, O., Niemist{\"o}, A.: A novel method for splitting
  clumps of convex objects incorporating image intensity and using rectangular
  window-based concavity point-pair search.
\newblock Pattern Recognition \textbf{46}(3), 741--751 (2013)

\bibitem{fernandez1995new}
Fern{\'a}ndez, G., Kunt, M., Zr{\"y}d, J.P.: A new plant cell image
  segmentation algorithm.
\newblock In: International Conference on Image Analysis and Processing, pp.
  229--234. Springer (1995)

\bibitem{gonzalez2014red}
Gonz{\'a}lez-Hidalgo, M., Guerrero-Pena, F., Herold-Garcia, S., Jaume-i
  Cap{\'o}, A., Marrero-Fern{\'a}ndez, P.D.: Red blood cell cluster separation
  from digital images for use in sickle cell disease.
\newblock IEEE journal of biomedical and health informatics \textbf{19}(4),
  1514--1525 (2014)

\bibitem{harris1988combined}
Harris, C.G., Stephens, M., et~al.: A combined corner and edge detector.
\newblock In: Alvey vision conference, vol.~15, pp. 10--5244. Citeseer (1988)

\bibitem{he2004curvature}
{He}, X.C., {Yung}, N.H.C.: Curvature scale space corner detector with adaptive
  threshold and dynamic region of support.
\newblock In: Proceedings of the 17th International Conference on Pattern
  Recognition, 2004. ICPR 2004., vol.~2, pp. 791--794 Vol.2 (2004).
\newblock \doi{10.1109/ICPR.2004.1334377}

\bibitem{he2014automated}
He, Y., Meng, Y., Gong, H., Chen, S., Zhang, B., Ding, W., Luo, Q., Li, A.: An
  automated three-dimensional detection and segmentation method for touching
  cells by integrating concave points clustering and random walker algorithm.
\newblock PloS one \textbf{9}(8) (2014)

\bibitem{kumar2006rule}
Kumar, S., Ong, S.H., Ranganath, S., Ong, T.C., Chew, F.T.: A rule-based
  approach for robust clump splitting.
\newblock Pattern Recognition \textbf{39}(6), 1088--1098 (2006)

\bibitem{latorre2013segmentation}
LaTorre, A., Alonso-Nanclares, L., Muelas, S., Pe{\~n}a, J., DeFelipe, J.:
  Segmentation of neuronal nuclei based on clump splitting and a two-step
  binarization of images.
\newblock Expert Systems with Applications \textbf{40}(16), 6521--6530 (2013)

\bibitem{matthews1975comparison}
Matthews, B.W.: Comparison of the predicted and observed secondary structure of
  t4 phage lysozyme.
\newblock Biochimica et Biophysica Acta (BBA)-Protein Structure
  \textbf{405}(2), 442--451 (1975)

\bibitem{mosaliganti2012acme}
Mosaliganti, K.R., Noche, R.R., Xiong, F., Swinburne, I.A., Megason, S.G.:
  Acme: automated cell morphology extractor for comprehensive reconstruction of
  cell membranes.
\newblock PLoS computational biology \textbf{8}(12), e1002780 (2012)

\bibitem{mosley2013balanced}
Mosley, L.: A balanced approach to the multi-class imbalance problem.
\newblock Ph.D. thesis, Iowa State University (2013)

\bibitem{otsu1979threshold}
Otsu, N.: A threshold selection method from gray-level histograms.
\newblock IEEE transactions on systems, man, and cybernetics \textbf{9}(1),
  62--66 (1979)

\bibitem{pavlidis1980algorithms}
Pavlidis, T.: Algorithms for shape analysis of contours and waveforms.
\newblock IEEE Transactions on pattern analysis and machine intelligence
  \textbf{1}(4), 301--312 (1980)

\bibitem{rodriguez2005new}
Rodr{\'\i}guez, R., Alarc{\'o}n, T.E., Pacheco, O.: A new strategy to obtain
  robust markers for blood vessels segmentation by using the watersheds method.
\newblock Computers in biology and medicine \textbf{35}(8), 665--686 (2005)

\bibitem{samma2010combining}
Samma, A.S.B., Talib, A.Z., Salam, R.A.: Combining boundary and skeleton
  information for convex and concave points detection.
\newblock In: 2010 Seventh International Conference on Computer Graphics,
  Imaging and Visualization, pp. 113--117. IEEE (2010)

\bibitem{song2009new}
Song, H., Wang, W.: A new separation algorithm for overlapping blood cells
  using shape analysis.
\newblock International Journal of Pattern Recognition and Artificial
  Intelligence \textbf{23}(04), 847--864 (2009)

\bibitem{wahlby2004combining}
W{\"a}hlby, C., Sintorn, I.M., Erlandsson, F., Borgefors, G., Bengtsson, E.:
  Combining intensity, edge and shape information for 2{D} and 3{D}
  segmentation of cell nuclei in tissue sections.
\newblock Journal of microscopy \textbf{215}(1), 67--76 (2004)

\bibitem{wang2012clump}
Wang, H., Zhang, H., Ray, N.: Clump splitting via bottleneck detection and
  shape classification.
\newblock Pattern Recognition \textbf{45}(7), 2780--2787 (2012)

\bibitem{wen2009delaunay}
Wen, Q., Chang, H., Parvin, B.: A delaunay triangulation approach for
  segmenting clumps of nuclei.
\newblock In: 2009 IEEE International Symposium on Biomedical Imaging: From
  Nano to Macro, pp. 9--12. IEEE (2009)

\bibitem{yan2011new}
Yan, L., Park, C.W., Lee, S.R., Lee, C.Y.: New separation algorithm for
  touching grain kernels based on contour segments and ellipse fitting.
\newblock Journal of Zhejiang University SCIENCE C \textbf{12}(1), 54--61
  (2011)

\bibitem{yeo1994clump}
Yeo, T., Jin, X., Ong, S., Sinniah, R., et~al.: Clump splitting through
  concavity analysis.
\newblock Pattern Recognition Letters \textbf{15}(10), 1013--1018 (1994)

\bibitem{zafari2015segmentation}
Zafari, S., Eerola, T., Sampo, J., K{\"a}lvi{\"a}inen, H., Haario, H.:
  Segmentation of partially overlapping nanoparticles using concave points.
\newblock In: International Symposium on Visual Computing, pp. 187--197.
  Springer (2015)

\bibitem{zafari2017comparison}
Zafari, S., Eerola, T., Sampo, J., K{\"a}lvi{\"a}inen, H., Haario, H.:
  Comparison of concave point detection methods for overlapping convex objects
  segmentation.
\newblock In: Scandinavian Conference on Image Analysis, pp. 245--256. Springer
  (2017)

\bibitem{zafari2020resolving}
Zafari, S., Murashkina, M., Eerola, T., Sampo, J., K{\"a}lvi{\"a}inen, H.,
  Haario, H.: Resolving overlapping convex objects in silhouette images by
  concavity analysis and gaussian process.
\newblock Journal of Visual Communication and Image Representation \textbf{73},
  102962 (2020)

\bibitem{zhang2020structure}
Zhang, Q., Wang, J., Liu, Z., Zhang, D.: A structure-aware splitting framework
  for separating cell clumps in biomedical images.
\newblock Signal Processing \textbf{168}, 107331 (2020)

\bibitem{zhang2017automated}
Zhang, W., Li, H.: Automated segmentation of overlapped nuclei using concave
  point detection and segment grouping.
\newblock Pattern Recognition \textbf{71}, 349--360 (2017)

\bibitem{zhang2012method}
Zhang, W.H., Jiang, X., Liu, Y.M.: A method for recognizing overlapping
  elliptical bubbles in bubble image.
\newblock Pattern Recognition Letters \textbf{33}(12), 1543--1548 (2012)

\end{thebibliography}

\end{document}